\documentclass[lettersize,journal]{IEEEtran}
\usepackage{amsmath,amsfonts}
\usepackage{algorithmic}
\usepackage{algorithm}
\usepackage{array}
\usepackage[caption=false,font=normalsize,labelfont=sf,textfont=sf]{subfig}
\usepackage{textcomp}
\usepackage{stfloats}
\usepackage{url}
\usepackage{verbatim}
\usepackage{graphicx}
\usepackage{cite}
\usepackage{xcolor}
\usepackage{lipsum}
\usepackage{multirow}
\usepackage{tabularx}
\usepackage{float}
\usepackage{orcidlink}
\usepackage{enumitem}
\usepackage{tabularx}

\usepackage{makecell}  % Package for wrapping text and centering in table cells

\usepackage[most]{tcolorbox}

\newtcolorbox{highlightedsubsection}{
  colback=yellow!10,    % background color
  colframe=black!50,    % border color
  boxrule=0.5pt,
  arc=3pt,
  left=6pt,
  right=6pt,
  top=6pt,
  bottom=6pt,
  enhanced,
}

\usepackage{soul}

\bstctlcite{IEEEexample:BSTcontrol}
\begin{document}

% Adjust cell padding and row height
\renewcommand{\arraystretch}{1.5}  % Adjust row height
\setlength{\tabcolsep}{5pt}  % Adjust horizontal padding

\title{The Role of Integrity Monitoring in Connected and Automated Vehicles: Current State-of-Practice and Future Directions}

\author{
    Saswat~Priyadarshi~Nayak$^{\orcidlink{0000-0003-2210-0021}}$,~\IEEEmembership{Student Member,~IEEE}, 
    and~Matthew~Barth$^{\orcidlink{0000-0002-4735-5859}}$,~\IEEEmembership{Fellow,~IEEE}%
    \thanks{Saswat Priyadarshi Nayak (Corresponding author) is with the Department of Electrical and Computer Engineering, University of California, Riverside, Riverside, USA (email: snaya004@ucr.edu).}
    \thanks{Matthew Barth is with the Department of Electrical and Computer Engineering, University of California, Riverside, Riverside, USA (email: barth@ece.ucr.edu).}
}


\maketitle

\begin{abstract}

Positioning integrity refers to the trust in the performance of a navigation system. Accurate and reliable position information is needed to meet the requirements of Connected and Automated Vehicle (CAV) applications, particularly in safety-critical scenarios. Receiver Autonomous Integrity Monitoring (RAIM) and its variants have been widely studied for Global Navigation Satellite System (GNSS)-based vehicle positioning, often fused with kinematic (e.g., Odometry) and perception sensors (e.g., camera). However, integrity monitoring (IM) for cooperative positioning solutions leveraging Vehicle-to-Everything (V2X) communication has received comparatively limited attention. This paper reviews existing research in the field of positioning IM and identifies various research gaps. Particular attention has been placed on identifying research that highlights
cooperative IM methods. It also examines key automotive safety standards and public V2X datasets to map current research priorities and uncover critical gaps. Finally, the paper outlines promising future directions, highlighting research topics aimed at advancing and benchmarking positioning integrity.

\end{abstract}


\section{Introduction}

Connected and Automated Vehicle (CAV) research is crucial to ensure safety, smooth traffic operation, and minimize energy consumption and carbon emissions~\cite{Guanetti2018ControlChallenges}. With the advent of sophisticated sensor platforms, computation hardware, and communication capability, it is now possible to deploy intelligent CAV applications through cooperation between various road agents. CAV applications can be broadly classified into three major use categories: safety, mobility, and environmental. Mobility and environmental applications such as eco-routing, eco-approach and departure (EAD) typically require coarse positioning accuracy (5-10 m) and/or lane level positioning accuracy ($< 1$ m), while safety-critical applications such as collision avoidance, autonomous intersection management, etc., generally require a where-in lane level accuracy ($<$ 0.2 m). Nigel et al. provided a qualitative analysis on various CAV applications and their positioning requirements in ~\cite{Williams2021AApplications}. The positioning accuracy required for various CAV applications can be achieved through onboard positioning sensors, such as the Global Navigation Satellite System (GNSS) receivers and Inertial Measurement Units (IMU), as well as perception sensors like cameras and LiDAR. Additionally, Roadside camera and LiDAR data can be broadcast over cellular or DSRC links and incorporated via relative-ranging techniques to enhance cooperative positioning~\cite{dePonteMuller2017SurveyVehicles}.

A CAV application's feasibility is dependent on the positioning performance of a specific sensor platform (standalone positioning) in the vehicle and/or its ability to use the information shared by neighboring road agents (cooperative positioning). Standalone positioning techniques include GNSS-based positioning~\cite{Groves2008PrinciplesApplications}, which is often fused with other onboard kinematic sensors such as wheel odometry~\cite{Liu2021Odometer-AidedCanyons}, Inertial Measurement Units (IMU)~\cite{Lim2018AugmentationArea}, and perception sensors such as radars, cameras, and LiDARs~\cite{Wen20223DCanyons}. On the other hand, cooperative positioning relies on relative ranging techniques and cooperative feature association techniques for vehicle localization.

The positioning performance of a navigation system is characterized by four key parameters: accuracy, availability, continuity, and integrity risk~\cite{Jing2022IntegrityChallenges}. For a given vehicular application, the system continuously monitors integrity risk, defined as the percentage of time it fails to raise an alert when the true position error exceeds the alert limit, by evaluating positioning accuracy in real time. If the system is unable to provide a reliable bound on the true position error over a predefined time window (i.e., a continuity breach), the integrity monitoring framework transitions to a fallback state. In simpler terms, the positioning solutions are made unavailable rather than being used for motion planning and control, which could be hazardous.

The research on integrity monitoring emerged alongside Global Positioning Satellite (GPS) technology in the field of aviation back in the 1980s, to provide a reliable navigation aid for aircraft. The Receiver Autonomous Integrity Monitoring (RAIM)~\cite{Pullen2020GNSSRAIM} algorithm was devised for single-fault detection using redundant pseudorange measurements from satellites of a single constellation, originally GPS. Over the years, other versions of RAIM, such as advanced RAIM (ARAIM), solution separation RAIM were developed for multi-constellation and multi-fault detection and isolation. However, RAIM cannot be directly applied to CAV applications because of its foundational assumptions, threat models, and operational constraints that are more suited for aviation. Unlike aviation, automotive navigation faces challenges such as multipath, non-line-of-sight, etc. Moreover, the positioning requirements in aviation are generally less stringent than those of safety-critical CAV applications. With the advent of multi-modal sensors and V2X information sharing, it is increasingly important to develop integrity monitoring frameworks that account for multi-sensor fusion and the performance characteristics of V2X communication.

However, in mixed-traffic scenarios, there is significant variation in vehicle types, sensor configurations, levels of cooperation, and positioning techniques. It is therefore unlikely that two vehicles equipped with the same sensors and computational hardware will exhibit identical navigation performance~\cite{Hassan2021ASystems}. Hence, it is essential to study and analyze the integrity of positioning systems under these varying conditions by surveying existing work and identifying key insights that can inform the development of integrity monitoring frameworks tailored to CAV-centric applications. The main contributions of this paper are:

\begin{enumerate}[wide, labelwidth=!, labelindent=0pt]

    \item Existing integrity monitoring work and key automotive safety standards are surveyed, covering both standalone and cooperative approaches.
    
    \item The main limitations in current IM research are identified and discussed, particularly in multisensor fusion, mixed traffic scenarios, and the lack of cooperative IM solutions.
    
    \item Emphasis is placed on developing cooperative integrity monitoring frameworks that leverage shared perception and V2X data in mixed-traffic environments.
    
    \item The paper provides a detailed discussion linking vehicular positioning uncertainties to Required Navigation Performance (RNP) metrics, and outlines key IM techniques including RAIM, Kalman filter residuals, model-based, and set-theoretic approaches.
    
    \item Existing V2X and cooperative datasets are reviewed to highlight their relevance and potential for benchmarking IM algorithms in various CAV applications.
    
    \item Finally, a clear set of research priorities and future directions is established for integrity monitoring research in the context of CAVs.
    
\end{enumerate}

Although mixed traffic in CAV applications includes other road users such as vulnerable road users (VRUs), this paper focuses exclusively on vehicle positioning. The different sections of this paper are organized as follows. Section~\ref{sec: CAV_uncertainties} discusses various uncertainties considered in past CAV research, i.e. model uncertainty, position uncertainty, etc. Integrity monitoring methods for vehicles via standalone and cooperative operations are presented in Section~\ref{sec: Pos_IM}. Section~\ref{sec: automotive_safety_standards} discusses key automotive safety standards in practice. Emerging V2X datasets 
are discussed in Section~\ref{sec: v2x datasets}. Section~\ref{sec:gaps_n_future_dir} explores existing research gaps and highlights potential opportunities for future work. The paper concludes with a comprehensive conclusion, synthesizing the key findings and implications of the research in Section~\ref{sec: conclusions}.

\section{Background: Consideration of Positioning Uncertainties in CAV Applications}
\label{sec: CAV_uncertainties}
Uncertainty originates from imperfect knowledge about the state or condition of a vehicle. Various navigation and control algorithms rely on mathematical models to define the vehicle's motion and interaction with their surroundings. These models are often an approximation of the real vehicle operations leading to model uncertainties~\cite{Li2023AnalysisUncertainties, Hu2022DistributedUncertainties}. In existing work, the vehicles are often assumed to be homogeneous across the same vehicle class, i.e. HDVs, AVs, CAVs. However, vehicles are different when it comes to car-following models or powertrain control strategies. Hence, the uncertainties of the model and the heterogeneity in mixed traffic should be taken into account when developing intelligent applications. Accurate state information, especially the vehicle's position, is critical in deciding whether an application could be deployed or not. Uncertainty in positioning information can sabotage driving functionalities and cause major safety concerns.

The position uncertainty of a vehicle can be attributed to the sensor suite available on the vehicle, along with any additional external sensor information that may aid the in-vehicle sensors. A fully connected and automated vehicle may turn itself into a degraded-CAV, AV only, CV only, and HDV while experiencing loss of communication and / or control. Hence, a sensitivity analysis is required to evaluate the positioning needs of a CAV application and to determine whether the necessary position accuracy can be achieved with just vehicular sensors or a combination of vehicular sensors with external sensor information~\cite{Williams2022PositionTraffic}. If where-in-lane-level position accuracy is ensured for vehicles, CAV applications such as Intersection Movement Assist (IMA) and Autonomous Intersection Management (AIM) can be implemented. If not, the application parameters may be dynamically adjusted so that the vehicle can perform less critical applications such as IMA or eco-approach and departure with basic longitudinal speed control. 

\begin{figure*}[ht]
        \centering
        \includegraphics[scale = 0.11]{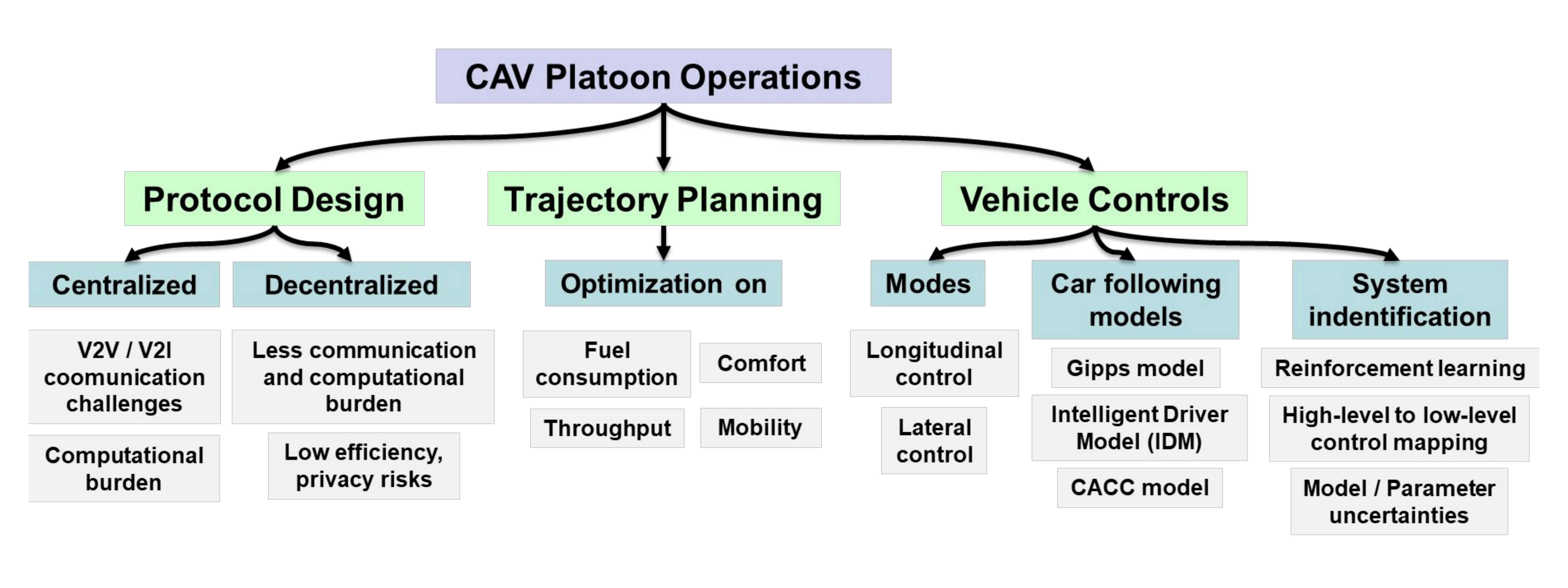}
        \caption{Fundamental operations of vehicle platooning, showcasing research opportunities and challenges in protocol design, trajectory planning, and vehicle controls.}
        \label{fig: platooning}
\end{figure*}

One of the promising CAV applications is some form of platooning~\cite{Li2022AOperations}. Platooning involves three fundamental maneuvers: merging, formation maintenance, and splitting. Merging and splitting involve coordinated actions between vehicles for safe lane changes, while formation maintenance involves longitudinal control along with an appropriate car-following model~\cite{Li2023AnalysisUncertainties}. A good deal of research has been done in this area, focusing mainly on protocol design and trajectory planning, as depicted in fig\ref{fig: platooning}. Li et al.~\cite{Li2020PlatoonExperiments} conducted experiments on platoon control with vehicles connected via V2X communication. A non-linear consensus-based longitudinal control was implemented considering communication probability in the car-following model. Luo et al. presented a cooperative control strategy to manage vehicle platoon in~\cite{Luo2023ModelingErrors} under measurement uncertainties. Real-world vehicle trajectories were taken from NGSIM dataset to simulate the experiments. Li et al.~\cite{Li2023AnalysisUncertainties} explored uncertainties in relative velocity and headway in analyzing car-following stability. Lu et al.~\cite{Lu2024UncertaintyAnalyses} proposed a stochastic model for mixed traffic modeling and considered CAV degradation under communication constraints. Furthermore, the efficiency and stability of the mixed traffic were evaluated under different CAV penetration rates. A distributed Model Predictive Control (MPC) technique was explored in~\cite{Hu2022DistributedUncertainties} for platoon control with model mismatch error and mixed disturbances. 

In addition to platooning experiments, other CAV applications, such as cooperative ramp merging and autonomous intersection management (AIM) have been studied in this context. Nigel et al.~\cite{Williams2022PositionTraffic} studied the merging of vehicles in a mixed-traffic scenario under position uncertainty. The impact of position uncertainty was studied in~\cite{Chamideh2023ImpactSystem} where a hierarchical MPC was developed to control the traffic and the vehicle at an intersection. Vitale et al.~\cite{Vitale2022AutonomousUncertainty,vitale_2023} proposed a mathematical framework for AIM problem considering uncertainty in location and provided an optimization technique to generate optimal vehicle acceleration profiles. Aoki and Rajkumar presented a data sharing framework~\cite{Aoki2022SafeVehicles} for intersection management utilizing a cooperative perception-based HD map.

\begin{figure}[ht]
        \centering
        \includegraphics[width=1\linewidth]{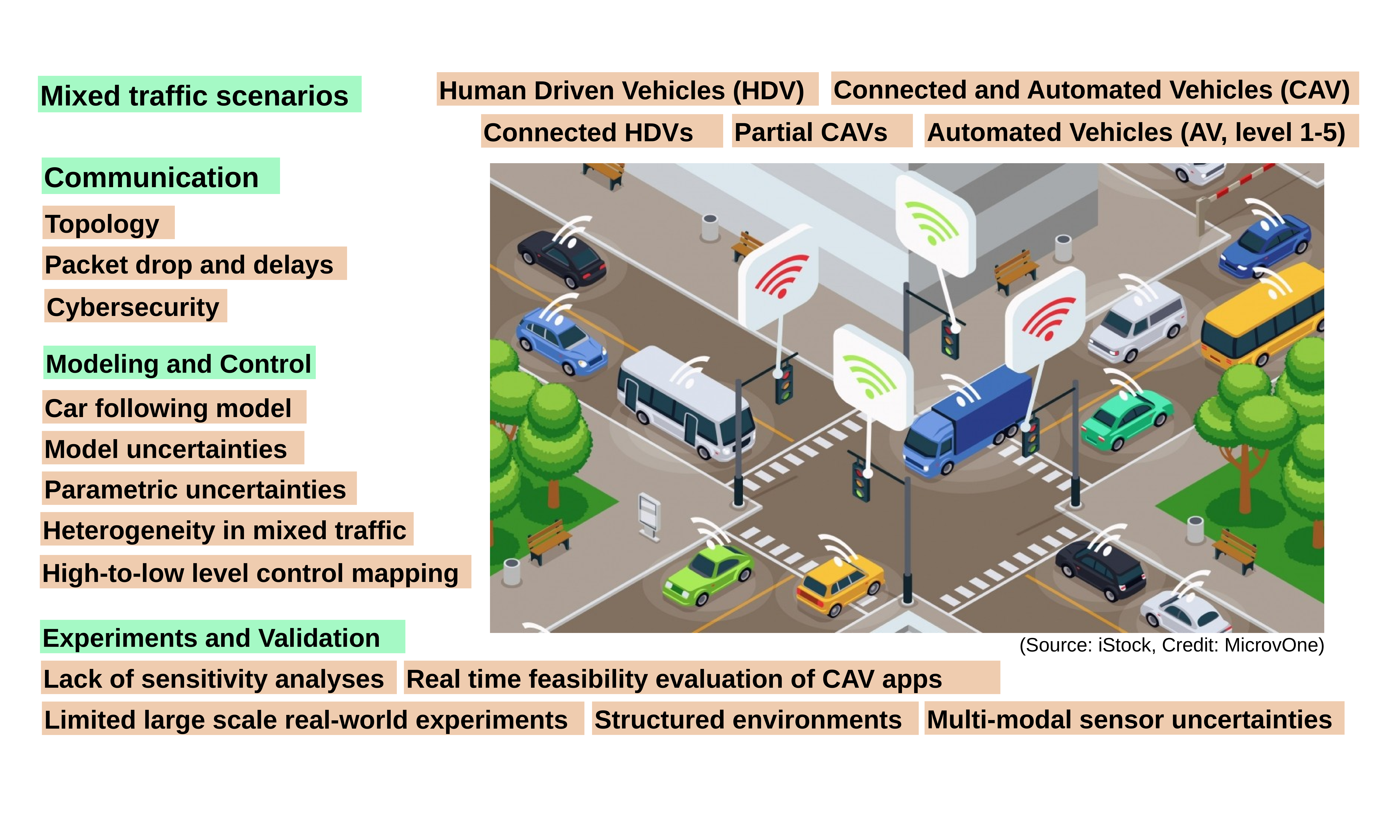}
        \caption{Identified research gaps in Connected and Autonomous Vehicle (CAV) applications. Gaps are marked in brown, while the corresponding research domains are highlighted in green.}
        \label{fig: CAV_research_gaps}
\end{figure}

Existing literature on the implementation and deployment of CAV applications focuses mostly on the planning aspect of the application. There is a lack of real-world experiments, as most of them are either conducted in simulation or controlled environments (see fig\ref{fig: CAV_research_gaps}). Multiple sensor modalities and their errors need to be quantified to properly evaluate their impact on the application's performance. Heterogeneity in car-following models and parameters needs to be considered, as existing research often assumes homogeneous vehicle types in a mixed traffic environment. Uncertainties in communication, measurements, and related processes must be studied to assess the integrity of vehicle navigation and to prepare CAVs for the road. Hence, this paper focuses on revisiting existing work on positioning integrity monitoring methods and identifying key gaps that will help improve modern navigation systems.

\section{Positioning Integrity Monitoring}
\label{sec: Pos_IM}

Integrity refers to ``trust'' in the performance of a system~\cite{Hassan2021ASystems}. In the context of Intelligent Transportation Systems (ITS), integrity has been evaluated using various strategies including Fault Detection and Isolation (FDI)~\cite{GNSS_FDE}, parity space~\cite{IM_parity_space}, Kalman filter residuals, cross-consistency checks~\cite{Balakrishnan2019AnLocalization}, context-aware monitoring~\cite{Context_aware_GPS_IM}, among others. A fault can occur at multiple levels, such as sensor level, signal transmission level, data processing level, system integration level, and application level~\cite{Raouf2022Sensor-BasedSurvey,solaas_anomaly,CostaDeOliveira2022RecentReview}. These multilevel faults are summarized in TABLE~\ref{tab: fault_levels}. Most of the FDI-based research has either been conducted for faults at the sensor (hardware) level or the application (software) level. Raouf et al. presented a sensor-based prognostic health management for vehicle ADAS\cite{Raouf2022Sensor-BasedSurvey}. This paper shall only focus on the system integration and application-level faults associated with vehicle positioning.

\begin{table}[!h]
    \caption{Fault Levels in Integrity Monitoring}
    \label{tab: fault_levels}
    \centering
    \begin{tabular}{|p{0.2\linewidth}|p{0.65\linewidth}|}
    \hline
    \textbf{Level} & \textbf{Fault}\\
    \hline
    Sensor & 
        \begin{itemize}
            \item Aging and faulty physical components
            \item Thermal, electromagnetic, harsh weather conditions
            \item Bias, scale factor errors, calibration errors
        \end{itemize} \\
    \hline
    Signal Transmission & 
        \begin{itemize}
            \item Electromagnetic interference, crosstalk, etc.
            \item Packet loss, transmission delay
            \item Time synchronization error
        \end{itemize} \\
    \hline
    Data Processing & 
        \begin{itemize}
            \item Faults in data filtering or fusion algorithms
            \item Software bugs, misinterpretation of data
            \item Incorrect system configurations
        \end{itemize} \\
    \hline
    System Integration & 
        \begin{itemize}
            \item Multipath error, reference signal errors
            \item Data fusion issues, sensor alignment and coordinate transformation errors
            \item Power overloading, memory constraints, cybersecurity, etc.
        \end{itemize} \\
    \hline
    Application & 
        \begin{itemize}
            \item Operational constraints, logic flaws, unoptimized application features
            \item Misinterpretation of sensor data
            \item Error handling issues, resource bottlenecks
        \end{itemize} \\
    \hline
    \end{tabular}
\end{table}

The integrity monitoring (IM) framework consists of a set of requirements for the navigation system that must be met to ensure safe operation. These requirements are subject to the positioning system used by the vehicle and the application it intends to perform. These requirements are referred to as Required Navigation Performance (RNP). The performance of the vehicle positioning and navigation system is quantified using four properties defined by the RNP framework - Accuracy, Availability, Continuity, and Integrity risk. These properties are quantified by four parameters~\cite{Worner2016IntegritySurvey,Tossaint2007TheAssessment}:

\begin{itemize}
    \item \textbf{Protection level (PL):} An upper bound on the position error (PE) with a specified confidence level. Protection levels can be defined for the longitudinal and lateral direction of a vehicle.
    \item \textbf{Alert limit (AL):} The maximum allowable error before an alert is triggered. AL is set by safety requirements and system specifications.
    \item \textbf{Time-to-alert (TTA):} The maximum allowable time from the occurrence of a fault to the time when an alert is issued.
    \item \textbf{Integrity risk (IR):} The likelihood of true position error exceeding the alert limit without raising an alert.
\end{itemize}

\begin{figure}[ht]
        \centering
        \includegraphics[width=1.1\linewidth]{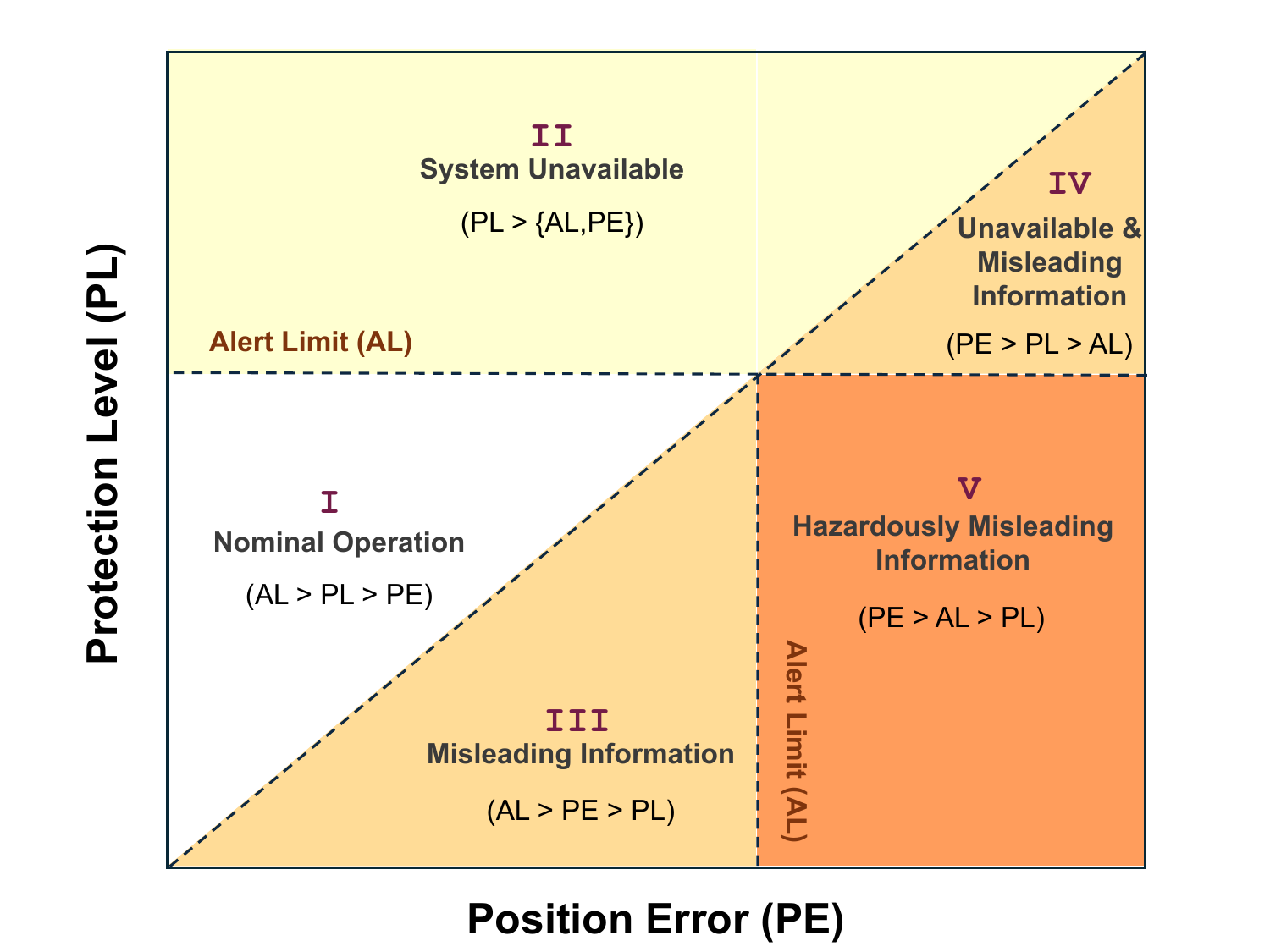}
        \caption{Stanford-ESA integrity diagram to monitor positioning integrity. The location of each (PL, PE) pair on the chart indicates whether the system is in nominal operation, unavailable, misleading information, or a hazardously misleading information zone.}
        \label{fig: ESA_diagram}
\end{figure}

The Stanford-ESA (Error Separation Algorithm) integrity diagram~\cite{Tossaint2007TheAssessment} is a graphical tool used to illustrate and assess the integrity of GNSS-based navigation systems, particularly in aviation. Although this plot was developed for GNSS positioning systems, it can be adapted and used for other navigation solutions. The PL estimated from standard deviations in position estimates and multipliers derived from desired confidence levels is plotted against the true position error over time across different scenarios. Different operating zones in the Stanford plot are discussed in detail in~\cite{Worner2016IntegritySurvey} and~\cite{Jing2022IntegrityChallenges}. Integrity risk is determined by the percentage of pairs (PL, PE) in the Hazardously Misleading Information (HMI) region without raising an alert, thus compromising vehicle safety. The different zones of the integrity diagrams, as shown in fig\ref{fig: ESA_diagram}, are discussed as follows:

\begin{enumerate}
    \item \textbf{Zone I - Nominal operation}: Both the protection level and the positioning error remain well within the alert limit, with the system conservatively estimating its error (i.e., PL $>$ PE). As a result, integrity is maintained, ensuring safe operation without the need for alerts.

    \item \textbf{Zone II - System Unavailable}: When the Protection Level (PL) exceeds the Alert Limit (AL), the uncertainty in the positioning estimate is too high for the application’s safety requirement. Even if the actual Position Error (PE) is below AL, the system can’t guarantee it — so it declares the service unavailable rather than risking unsafe operation. For example, GPS-only positioning in urban environments can have errors of 5–10 meters due to multipath and signal blockage, making it unreliable for lane-centering control in automated vehicles, which typically requires errors within 1 meter.

    \item \textbf{Zone III - Misleading information}: When AL $>$ PE $>$ PL, the system underestimates the actual position error, leading to overconfidence. For example, in infrastructure LiDAR-based vehicle tracking, the lateral protection level (PL) might be estimated at 0.4 meters under normal conditions. However, due to partial occlusion or a sparse point cloud — such as when a vehicle is partially blocked by another object — the actual lateral position error (PE) may jump to 1.0 meter. If the alert limit (AL) is 1.5 meters (i.e., half the lane width), the system still considers the output safe. Yet, this unnoticed degradation poses a risk, as the error is significantly higher than the system claims, reducing the effective safety margin and leading to misleading information.

    \item \textbf{Zone IV - Unavailable and misleading}: Under this condition, the system not only fails to meet the required accuracy (PL $>$ AL), but also underestimates how bad the error is (PE $>$ PL). This is a critical situation as the system is both unavailable and misleading.

    \item \textbf{Zone V - Hazardously misleading information}: Hazardously Misleading Information (HMI) occurs when the system underestimates its actual positioning error, resulting in false confidence. For example, in lane-level positioning, the vehicle may have already drifted out of its lane (PE $>$ AL), but the system reports a protection level (PL) well within the alert limit (PL $<$ AL). Since no alert is triggered, the system continues to operate under the assumption that the position is safe. This undetected failure poses a serious risk in safety-critical applications. 
\end{enumerate}

\begin{figure*}[ht]
        \centering        
        \includegraphics[width=0.9\textwidth]{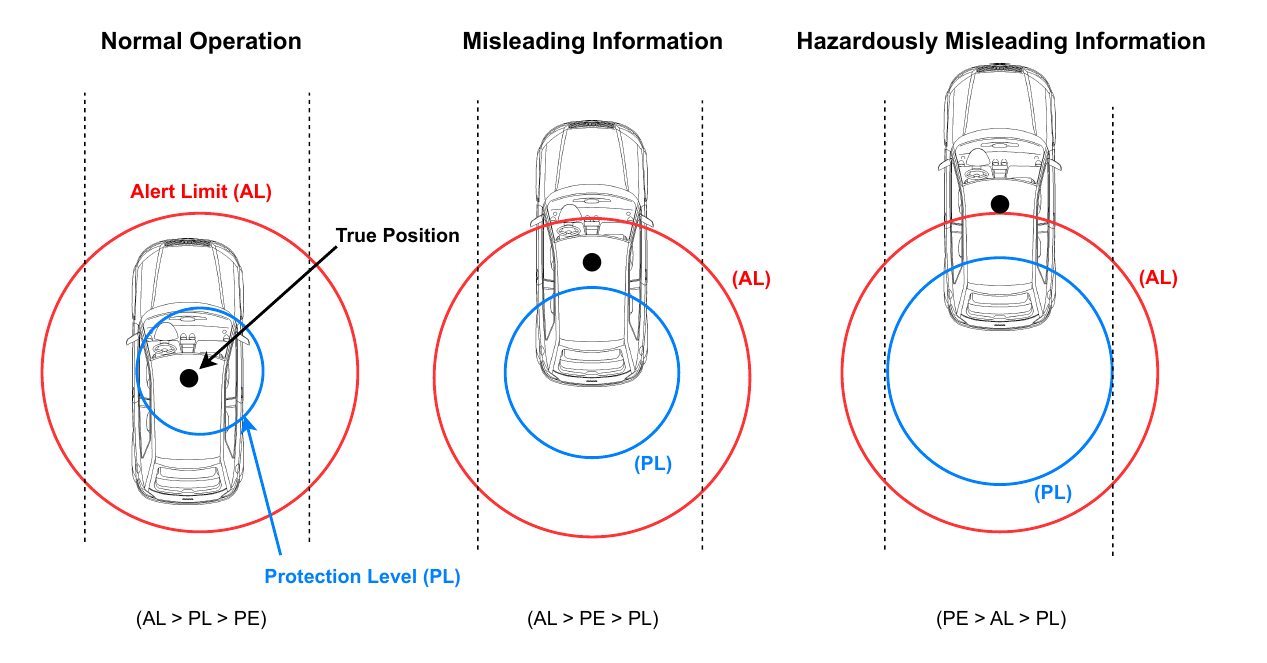}
        \caption{Vehicle’s positioning performance transitioning from nominal operation to misleading information and finally to hazardously misleading information. During the misleading stage, the system exhibits false confidence despite increased error. In the hazardously misleading stage, it fails to trigger an alert as the protection level remains below the alert limit.}
        \label{fig: increasing_PE}
\end{figure*}

Fig\ref{fig: increasing_PE} illustrates how a vehicle’s positioning degrades from normal operation into an integrity-risk state. This shift can result from erroneous measurements—such as GNSS signal blockage or multipath errors—that undermine the position-estimation process. For this reason, GNSS is commonly fused with onboard sensors or supplemented by cooperative V2X information.

\subsection{\textbf{Overview of Key Integrity Monitoring Techniques}}

Integrity monitoring ensures that the estimated position from a localization system remains reliable, particularly in safety-critical applications such as autonomous driving and aviation. The goal is to detect faults or inconsistencies in sensor data or system behavior and to verify whether the estimated state stays within acceptable limits. This section outlines four widely used integrity monitoring techniques: Receiver Autonomous Integrity Monitoring (RAIM), originally developed for GNSS; Kalman Filter-based methods, which extend RAIM to dynamic systems; model or coherence-based approaches that check consistency across independent sensor sources; and set-theoretic methods, such as interval analysis, which rely on bounded-error assumptions rather than probabilistic models (. Each method is suited to different use cases and offers specific strengths depending on system design and sensor characteristics.

\begin{enumerate}[leftmargin=*, labelindent=0pt]

    \item \textbf{Receiver Autonomous Integrity Monitoring (RAIM):}\\
    RAIM checks the consistency of pseudorange measurements received from multiple satellites by evaluating the residuals between observed and expected values. Classical RAIM, developed in the 1980s, is designed to detect a single satellite fault within a single GNSS constellation, originally GPS. In contrast, Advanced RAIM (ARAIM) extends this capability to handle multiple faults and supports multi-constellation setups. For a basic 3D position fix, signals from at least four satellites are required, while implementing RAIM requires at least five. The expected pseudoranges are computed based on the receiver’s estimated position and satellite geometry; any significant discrepancy between these and the actual measurements results in large residuals, signaling a potential fault. When such faults are detected, the system raises an integrity alert to either isolate or exclude the affected satellite’s data from the position solution.

    The least squares (LS) and weighted least squares (WLS) methods are two of the popular methods to compute residuals in pseudorange measurements and detect faults. The linearized GNSS pseudorange observation equation is formulated as:
    \begin{equation}
        y = Hx + \epsilon
    \end{equation}
    , where y is an nx1 vector containing actual pseudorange measurements minus expected range measurements based on the location of the (n) satellites and estimated receiver position, x is an unknown 4x1 vector including receiver position and clock bias; H is an nx4 observation matrix; $\epsilon$ is an n-dimensional column vector storing measurement noises; W is the weighted matrix represented by, \(\mathbf{W} = \mathrm{diag}\left(\frac{1}{\sigma_1^2}, \frac{1}{\sigma_2^2}, \ldots, \frac{1}{\sigma_n^2}\right)\). The estimated state vector and pseudorange residuals are computed as follows,
    \begin{equation}
        x_{wls} = (H^TWH)^{-1}H^TWy
    \end{equation}

    \begin{equation}
        w = y - Hx_{wls}
    \end{equation}

    The sum of squared errors could be calculated from the residuals by,
    \begin{equation}
        SSE = w^Tw
    \end{equation}

    The test statistic ($T_s$) used for fault detection could be computed using,
    \begin{equation}
        T_s = \sqrt{SSE/(n-4)}
    \end{equation}

    The test statistics follows a chi-square distribution with (n-4) degrees of freedom. Given the probability of false alarms ($P_{fa}$) the test threshold (T) can be computed using eq.~\ref{eq:sse} and eq.~\ref{eq:T},
    \begin{equation}
    \label{eq:sse}
        P(SSE < t) = 1 - P_{fa}
    \end{equation}
    \begin{equation}
    \label{eq:T}
        T = \sqrt{t/(n-4)}
    \end{equation}
    If the test statistic ($T_s$) is less than the threshold ($T$), no fault is detected. However, if $T_s > T$, a fault is present and must be identified using appropriate fault isolation techniques. Apart from least squares techniques, there exist other fault detection (FD) techniques used by other variants of RAIM algorithm. These include solution separation (SS), multi-hypothesis solution separation, chi-square method, ambiguity resolution algorithm (LAMBDA) and many more. Readers are encouraged to refer to surveys~\cite{Pullen2020GNSSRAIM, Zhu2018GNSSLiterature, IM_GNSS_RAIM} to gather additional information on advanced RAIM (ARAIM). relative RAIM, extended RAIM, carrier-based RAIM, time RAIM, vision-aided RAIM.\\
    
    \item \textbf{Kalman Filter Residual-based IM:}\\
    Kalman filter is a special case of least-squares estimators, specifically, an optimal, recursive, weighted least squares estimator designed for dynamic systems under Gaussian noise assumptions. At each time instance $t_k$, the innovation or residual $r_k$ is calculated as the difference between the actual measurement $z_k$ and the predicted measurement $\hat{z}_k = H\hat{x}_{k|k-1}$,
    \begin{equation}
        r_k = z_k - H\hat{x}_{k|k-1}
    \end{equation}
    , where $\hat{x}_{k|k-1}$ is the predicted state and H is the observation matrix. The covariance of the residual is computed as,
    \begin{equation}
        S_k = H_{k}P_{k|k-1}H_{k}^T + R_k
    \end{equation}
    with $P_{k|k-1}$ is the predicted error covariance and $R_k$ is the measurement error covariance. A chi-squared test statistic is defined as,
    \begin{equation}
        T_k = r^T_{k}S^{-1}_{k}r_k
    \end{equation}
    Similar to the least squares method used in RAIM, a fault is declared if the test statistics $T_k$ exceed the test threshold corresponding to a given false alarm probability. While traditional RAIM methods rely on least squares estimation, Kalman filter-based residuals have been explored as an alternative for fault detection in multi-sensor fusion systems combining GNSS with sensors such as cameras and LiDAR~\cite{Raouf2022Sensor-BasedSurvey}. For non-linear systems, Variants of the Kalman filter such as Extended KF, Unscented KF could be used in~\cite{Gottschalg2021ComparisonDriving, Ansari2015ADSRC}. \\

    \begin{figure*}[ht]
        \centering        
        \includegraphics[width=\textwidth]{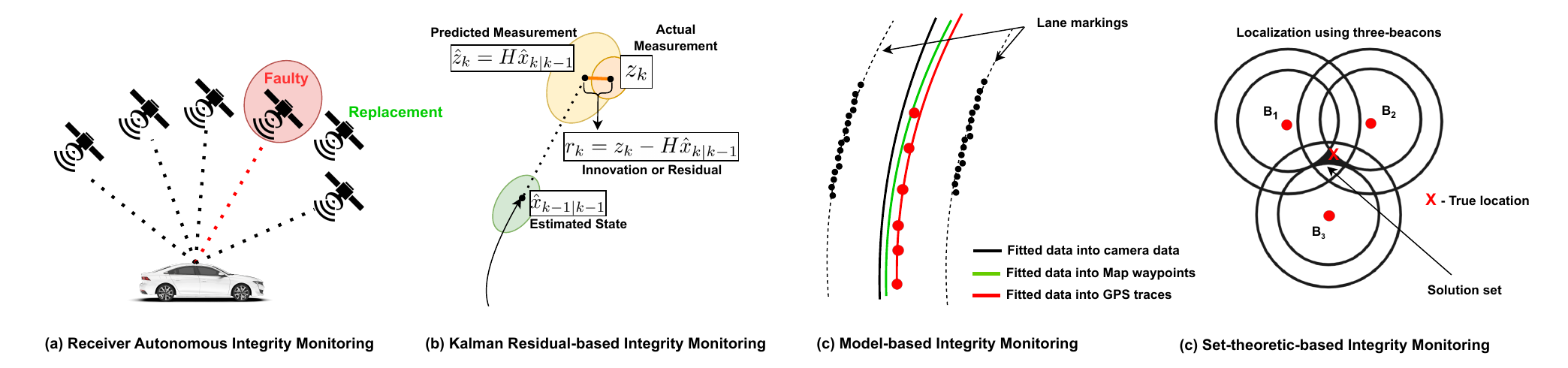}
        \caption{Schematic overview of four integrity-monitoring approaches:
        (a) Receiver Autonomous Integrity Monitoring (RAIM) evaluates pseudorange residuals to detect satellite faults;
        (b) Kalman filter residual–based IM calculates the innovation for filter consistency;
        (c) Model-based IM performs cross-consistency checks by fitting environmental features (map waypoints, GPS traces, camera-detected lane markings) across multiple sensors; and
        (d) Set-theoretic IM uses interval-based methods to compute feasible solution sets that provide a guaranteed bound for the true vehicle position.}
        \label{fig: key_im_methods}
    \end{figure*}

    \item \textbf{Model \& Coherence based IM:}\\
    Localization errors in perception sensors can arise not only at the system or application level but also at the signal or data level~\cite{Zhu2022IntegrityProspects}. Therefore, it is important to cross-check data consistency at the sensor source to prevent the propagation of errors into downstream modules. The key idea behind model consistency or coherence estimation across multiple sensor data streams is to assess the similarity in their perception of external environmental features. These features may include lane markings, curbs, landmarks, or other identifiable markers. In~\cite{Balakrishnan2019AnLocalization, Balakrishnan2020IntegrityLocalization}, a cross-consistency check was performed to evaluate Goodness-of-Fit (GoF) among data sources.

   Given \( N \) spatial sensor sources denoted as \( \mathcal{S} = \{S_1, S_2, \ldots, S_N\} \), each providing 2D data points for environmental features (curb, lanes, etc.) in a common reference frame (e.g., ego-frame). \( \mathbf{D}_i = \{(x_{i1}, y_{i1}), \ldots, (x_{iM_i}, y_{iM_i})\} \subset \mathbb{R}^2 \) -  represents spatial data from the sensor \( S_i \). The common feature (e.g., lane marking) is fitted with a second-degree polynomial model:
    
    \begin{equation}
         p_i(x) = a_i x^2 + b_i x + c_i
    \end{equation}
    
    , where $p_i$ represents the model and ${a_i,b_i,c_i}$ represent the polynomial coefficients estimated via robust regression. To evaluate the model \( p_i \) on data from sensor \( S_j \), a cross-model polynomial is defined:
    \begin{equation}
      p_{ij}(x) = a_i x^2 + b_i x + c_j   
    \end{equation}
   
    , where \( c_j \) is taken from \( p_j \) to align the offset to match the data from $S_j$. The curvature at point \( x \) for model \( p_{ij} \) is defined as:
    \begin{equation}
    \kappa_{ij}(x) = \frac{2a_i}{\left(4a_i^2 x^2 + 4a_i b_i x + b_i^2 + 1\right)^{3/2}}
    \end{equation}
    
    The computed curvature is used to weight the residuals. The weighted residual error for model \( p_{ij} \) on data \( \mathbf{D}_j \) is computed as:
    \begin{equation}
    e_{ij} = \frac{1}{M_j} \sum_{k=1}^{M_j} \kappa_{ij}(x_{jk}) \cdot \left| y_{jk} - p_{ij}(x_{jk}) \right|
    \end{equation}

    The error $e_{ij}$ is the GoF marker and quantifies the disagreement of sensor $i$'s model with data from sensor $j$. All pairwise cross-consistency values are collected in matrix \( E \in \mathbb{R}^{N \times N} \):\[
    E = [e_{ij}], \quad i,j = 1, \ldots, N
    \]
    
    Integrity markers \( W_i \) for each sensor \( S_i \) is defined as:
    \begin{equation}
    W_i = 1 - \frac{\sum_{j=1}^{N} e_{ij}}{\sum_{i=1}^{N} \sum_{j=1}^{N} e_{ij}},
    \end{equation}
    , such that $\sum_{i=1}^{N} W_i = 1$. A greater agreement with other sensor sources leads to a higher value of $W_i$. Any discrepancy or fault in sensor information would lead to a higher residual and a lower integrity score. \\

    \item \textbf{Set Theoretic IM:}\\
    The aforementioned IM techniques belong to fault detection (FD) methods, which rely on probabilistic models to compute the protection level (PL). However, the bounded error (BE) approaches provide a deterministic solution on the error bound and thus can work with unknown or non-gaussian noise distributions~\cite{Chachuat2015Set-TheoreticSystems,Su2022AdvancesNavigation}. It also guarantees the containment of measurements within a bound often represented using sets, which is critical in unpredictable and safety-critical scenarios involving autonomous driving~\cite{Maharmeh2025ASystems}. 
    
    Bounded error approaches include interval analysis, zonotope estimation, set-membership estimation, set-inversion via interval analysis (SIVIA), etc. Among these, interval analysis is a set-theoretic technique where uncertainties are modeled as closed intervals rather than probabilistic distributions~\cite{Worner2016IntegritySurvey}. Each scalar variable \( x \) is represented as an interval, $[x] = [x_{min}, x_{max}]$, 
    and a vector of variables \( \mathbf{x} \in \mathbb{R}^n \) is described using a n-dimensional box, which is the Cartesian product of scalar intervals:
    \begin{equation}
    [\mathbf{x}] = [x_1] \times [x_2] \times \dots \times [x_n].
    \end{equation}
    
     A scalar function \( f(x) \) can be extended to an interval function \( [f]([x]) \), which evaluates as:
    \begin{equation}
        [f]([x]) = \{ f(x) \mid x \in [x] \}.
    \end{equation}
    
    An inclusion function is one that always contains the true image of the function, meaning:
    \begin{equation}
    f([x]) \subseteq [f]([x]).
    \end{equation}
    In most practical cases, inclusion functions over-approximate the actual range due to variable dependencies or nonlinearity, but they ensure that the true value is always enclosed. In the context of estimation, interval analysis can be used to solve constraint satisfaction problems (CSPs) where the objective is to find the smallest box \( [x] \) such that all constraint functions \( f_i(x) \in [z_i] \) are satisfied. When some measurements are suspected to be outliers, a \( q \)-relaxed CSP can be formulated, which allows up to \( q \) constraints to be violated. The integrity risk associated with this relaxation can be expressed as:
    \begin{equation}
    R = 1 - \sum_{k = m - q}^{m} \binom{m}{k} p^k (1 - p)^{m - k},
    \end{equation}
    where \( m \) is the total number of constraints, \( q \) is the maximum number of tolerated outliers, and \( p \) is the probability that a measurement lies within its bounded interval. The method is particularly useful for integrity monitoring in localization tasks, as it provides hard guarantees even under bounded noise and measurement faults.
   
\end{enumerate}

\subsection{\textbf{In-vehicle (Standalone) Integrity Monitoring}}
\label{subsec: standalone_IM}

FDI-based integrity monitoring (IM) methods such as Receiver Autonomous Integrity Monitoring (RAIM) and its variants have been extensively used in the Global Navigation Satellite System (GNSS)-based systems to ensure the accuracy and reliability of position data provided by the receiver \cite{IM_GNSS_RAIM}. In the literature, different integration schemes and IM methods have been explored to fuse information from GNSS receivers, the inertial measurement unit (IMU), and vehicle odometry \cite{Jing2022IntegrityChallenges}. 

Automated vehicle research and their applications have surged in the last decade with the advent of advanced perception sensors and improved onboard computational capabilities. Reference \cite{Worner2016IntegritySurvey} surveys integrity for autonomous driving and presents two approaches: (1) FDI, and (2) bounded-error assumption to evaluate localization safety in autonomous vehicles. Apart from the widely used FDI methods, other IM methods have been developed for various driving scenarios and ITS applications. Integrity methods for sensor data fusion algorithms were developed in \cite{Gottschalg2021ComparisonDriving} to estimate the vehicle's dynamic state during automated driving. In this work, Kalman Integrated Protection Level (KIPL) performed the best while estimating protection level (PL) compared to the kSigma and Advanced-RAIM (ARAIM) method~\cite{Gottschalg2020IntegrityDriving}. Cross-consistency metric between multi-modal data sources is proposed in \cite{Balakrishnan2020IntegrityLocalization} instead of classical FDI schemes to evaluate localization integrity in automated vehicles. 

Visual information for navigation has been a well-researched topic since the last decade. Direct, indirect, and deep learning-based approaches have been used for relative pose estimation. In most research, robustness has been a key criterion for evaluating visual-navigation methods. Although state-of-the-art approaches may excel in information availability and continuity, they often fail to meet the minimum integrity requirements. Integrity frameworks define additional parameters such as accuracy and risk, for difficult navigation scenarios where reliable estimates are more crucial than outputting the best available solution under high risk. 

References \cite{Mario2010IntegrityMO} and~\cite{AlHage2019HighMeasurements} propose integrity monitoring for camera-based lane marking detection and removal of faulty measurements by monitoring the residuals of the estimated position. Visual information has also been used to aid existing GNSS measurements and help identify and eliminate Non-Line-of-Sight (NLOS) errors in deep urban canyons \cite{Bai2020UsingCanyons}. Zhu and Taylor \cite{Zhu2017ConservativeNavigation} studied the effect of correlated measurements on integrity in visual navigation. Zhu et al. proposed a technique to quantify the feature association error in \cite{zhu2019cooperative}. The research so far has focused on a specific type of error associated with visual navigation. The errors are often not modeled correctly and there is a lack of integrity monitoring methods for camera-only solutions. In \cite{Zhu2022IntegrityProspects}, a preliminary integrity monitoring framework is presented for a feature-based relative pose estimation problem in visual navigation. Additionally, the authors point out the challenges and developments in integrity analysis for visual navigation.

\begin{table*}[!ht]
\centering
\caption{Standalone and Cooperative Integrity Monitoring Techniques}
\label{tab: IM methods}
\begin{tabular}{|p{2cm}|p{0.6cm}|p{1.7cm}|p{3cm}|p{2.7cm}|p{6cm}|}
\hline
\textbf{Paper} & \textbf{Year} & \textbf{Mode} & \textbf{Data Source} & \textbf{Techniques} & \textbf{Key Takeaways (+) \& Identified Gaps (-)}\\
\hline
 Hassan et al. \cite{Hassan2021ASystems} & 2021 & Standalone & GNSS, IMU, Odometry, LiDAR & Least-squares-based ARAIM & \textbf{(+) :} Suitable for loosely coupled integration with other sensors and multi-faul multi-constellation scenarios  \newline \textbf{(-) :} Need for IM architectures for diverse non-GNSS sensors or positioning solutions\\
\hline
Worner et al. \cite{Worner2016IntegritySurvey} & 2016 & Standalone & N.A & BE - Interval analysis  &  \textbf{(+) :} Guaranteed bound on PE, Robust to outliers \newline \textbf{(-) :} Conservative, Computationally intensive, scalability issue for large state space\\
\hline
Jing et al. \cite{Jing2022IntegrityChallenges} & 2022 & Standalone \& Cooperative & GNSS/IMU, Map, Wireless & RAIM  &  \textbf{(+) :} Classical RAIM is easy to implement and suitable for real-time operations \newline \textbf{(-) :} Lack of comparable protection level (PL) computation from existing IM methods under the same vehicle testing environment \\
\hline
Raouf et al. \cite{Raouf2022Sensor-BasedSurvey} & 2022 & Standalone & GNSS, Camera, LiDAR, Radar & KF-based RAIM  &  \textbf{(+) :} Fault identification and diagnosis for overall sensors of ADAS for AVs \newline \textbf{(-) :} Need for comprehensive prognostic health management considering variation in vehicle type, sensor modality \& algorithms\\
\hline
Balakrishnan et al. \cite{Balakrishnan2019AnLocalization, Balakrishnan2020IntegrityLocalization} & 2020 & Standalone & GPS, Camera, LiDAR, Map & Model/Coherence-based IM  &  \textbf{(+) :} Robust against perception errors- partial detections, occlusions \newline \textbf{(-) : } Inability to detect common features in multi-modal sensor data\\
\hline
Gottschalg et al. \cite{Gottschalg2021ComparisonDriving, Gottschalg2020IntegrityDriving} & 2021 & Standalone & GNSS, IMU, Odometry & kSigma, KIPL, ARAIM  &  \textbf{(+) :} KIPL outperforms other methods in terms of accuracy and availability \newline \textbf{(-) :} Integrity algorithms yield varying protection levels, affecting availability and requiring selection based on application\\
\hline
Ansari et al. \cite{Ansari2015ADSRC} & 2015 & Cooperative & GNSS/RTK, IMU, DSRC & PRMF, RAIM, KF-based fault detection  &  \textbf{(+) :} Real-time IM benchmarking with DSRC availability monitoring \newline \textbf{(-) :} BSM reception affected by crowded/saturated network condition\\
\hline
Ansari \cite{Ansari2020CooperativePositioning} & 2020 & Cooperative & GNSS, DSRC & KF, EKF, PF, MR, MFNN  &  \textbf{(+) :} Motion Model and Maneuver
Recognition (MR) outperformed other methods in accuracy and computation complexity  \newline \textbf{(-) :} Discussed motion prediction models lack precision for advanced safety tasks \\
\hline
Liu et al. \cite{Liu2020AGNSS} & 2020 & Cooperative & GNSS, DSRC, Road map & Hybrid-RAIM, Virtual satellite measurements  &  \textbf{(+) :} Overcomes limited GPS coverage constraints under RAIM unavailability \newline\textbf{(-) :} Doesn't outperform conventional RAIM in open-sky condition\\
\hline
Maaref et al. \cite{Maaref2022AutonomousIMUb} & 2022 & Cooperative & GNSS, IMU, Cellular-LTE & SOP-based RAIM  &   \textbf{(+) :} Cellular pseudoranges aiding for IMU in the absence of GNSS signals (Urban canyons)  \newline \textbf{(-) :} Prior knowledge of cellular tower position, more multipath error compared to GNSS, estimation of cell-tower clock bias\\
\hline
Xiong et al. \cite{Xiong2021IntegrityPositioning} & 2021 & Cooperative & GNSS, UWB & Kalman residual-based CIM  &   \textbf{(+) :} High fault detection sensitivity compared to RAIM, fault detection and diagnosis in cooperative positioning \newline \textbf{(-) :} UWB ranging affected by Non-Line-of-Sight (NLOS) errors, Needs multi-sensor testing and validation \\
\hline
Schon et al. \cite{Schon2018IntegrityNetworks} & 2018 & Cooperative & GNSS, IMU, stereo camera, laser scanner & Interval set theory  &  \textbf{(+) :} Boundedness of estimates \& time-synchronization error in multi-sensor system \newline \textbf{(-) : }Overestimation of error bounds\\
\hline
\end{tabular}
\end{table*}

\subsection{\textbf{Cooperative Integrity Monitoring}}
\label{subsec: cooperative_IM}

IM methods have been extensively studied and developed for in-vehicle sensor systems primarily consisting of GNSS receivers often integrated with IMUs, vehicle odometry, and perception sensors such as radar, camera, and LiDAR. The localization performance and safety of a vehicle have mainly been studied from the vehicle's perspective focusing on enabling automated driving functions through onboard sensors and computing platform. 

AVs constitute a small percentage of traffic on the road today. There is a surge in V2V/V2I/V2P/V2X research focusing on information sharing between road agents for improved positioning, navigation, and control. Given the number of sensor sources and the amount of data shared between road agents, it is imperative to develop integrity monitoring frameworks for cooperative scenarios. Despite the growing interest in this area, there is a noticeable gap in the current literature that addresses the ``cooperative-IM" framework. For CAV applications, only a few studies have thoroughly examined different ways to evaluate integrity risks in wireless information exchange, indicating a need for further research to propose new RNP parameters~\cite{Jing2022IntegrityChallenges}.

In addition to the standard parameters used in the traditional IM framework, i.e. accuracy, availability, continuity, integrity, two other parameters - timeliness \& interoperability, have been proposed in the literature to evaluate the performance of cooperative-ITS (C-ITS) applications~\cite{Green2013VehicleDocument}. In~\cite{Ansari2015ADSRC}, a runtime IM framework for real-time relative positioning was introduced focusing on sharing raw GPS observation data through Dedicated Short Range Communications (DSRC). The author of previous work also explored a real-time relative position prediction strategy~\cite{Ansari2020CooperativePositioning} in a V2V environment using several position prediction techniques to compensate for the loss of the V2V DSRC link. In this work, only road-level accuracy ($<$5 m error) could be guaranteed. A hybrid integrity monitoring method is proposed in~\cite{Liu2020AGNSS} that overcomes the limitations of conventional RAIM technique under limited GNSS satellite coverage. It generates virtual satellite measurements using DSRC range rate measurements from neighboring vehicles and road map data. Recently, signals of opportunity (SOPs) such as AM/FM radio, cellular signals, etc., have aided the IMU in the absence of GNSS signals. Cellular signals in particular are of great interest due to their availability in urban canyons. A SOP-based RAIM framework was developed in~\cite{Maaref2022AutonomousIMUb} which fuses cellular pseduorange measurements with IMU measurements. The proposed method showed a reduction of position Root Mean Square Error (RMSE) by 66\%.

In existing studies, the IM framework majorly focuses on improving GNSS-based positioning systems either through an onboard IMU, vehicle odometry, or use of external information like SOPs, DSRC ranging, pseudorange differences relayed through wireless communication. IM methods have been developed for perception sensors such as cameras, LiDARs, radars, etc., but as a part of the onboard or in-vehicle perception system. The sharing of perception data in raw or processed format via V2V/V2I means has proven to enhance the detection of surrounding vehicles and improve localization performance~\cite{Bai2022CyberLiDAR,Nayak2023EvaluationModels, Nayak2024Infrastructure-AssistedMeasurements}. However, research on cooperative integrity monitoring for sharing perception data through wireless communication has been limited. Xiong et al. experimented with a cooperative integrity monitoring (CIM) method with FDE based on the innovation residual of the Kalman filter~\cite{Xiong2021IntegrityPositioning}. The CIM method was better at fault detection in GNSS and ultra-wide band (UWB) measurements compared to the traditional RAIM method \& CERIM method~\cite{Rife2011CollaborationEnhancedRI}. Various research experiments on integrity monitoring for automotive sensor fusion algorithms have been presented in~\cite{Schon2018IntegrityNetworks}. In this work, the authors have proposed integrity concepts based on interval set theory. A dynamic sensor network is created consisting of three vehicles equipped with stereo cameras, laser scanners, GNSS receivers, and IMUs. Issues such as inconsistency in GNSS measurements, random and bounded uncertainties are addressed in this paper.

The development of cooperative integrity monitoring methods is still in its early stages. The question remains whether the empirically determined Required Navigation Performance (RNP) values hold consistent across different driving conditions, positioning hardware, and communication topologies. Inconsistencies in RNP parameters in CAV applications would lead to multiple solutions from various automotive OEMs resulting in loss of standardization. In \cite{Feng2018DeterminationApplications}, the authors have defined the minimum operational performance standards that can be adjusted to meet the needs of a specific safety application based on the positioning capabilities of a vehicle. The ESA-funded P-CAR project~\cite{Valentini2022OnInfrastructure} focuses on identifying and validating connected and automated driving (CAD) functions to meet safety integrity needs. It aims at creating realistic Hardware-in-Loop (HiL) simulations and appropriate fault injection strategies to evaluate events with integrity risks that might be rare and difficult to test in real-world conditions. Based on the initial stakeholder surveys, truck platooning is being favored to test the efficacy of V2V communication, whereas, applications such as highway pilot, urban automated shuttle, and valet parking are key contenders in the navigation segment. Various cooperative integrity monitoring strategies and key findings are summarized in TABLE~\ref{tab: IM methods}. While aspects such as computational demand and latency are important for deployment, a quantitative, apples-to-apples comparison would require a dedicated meta-analysis or controlled simulation studies across diverse positioning techniques, sensor modalities, cooperation level (V2V, V2I, etc.), and environmental conditions (open-sky, highway, tunnel), which is beyond the scope of this paper.

\section{Automotive Safety Standards and Protocols}
\label{sec: automotive_safety_standards}

Vehicles depend on several automotive safety standards that are essential for precise navigation, effective collision avoidance, and ensuring the overall safety and reliability of the system. These safety standards are defined and maintained by various organizations around the world such as the Society of Automotive Engineers (SAE), the International Organization for Standardization (ISO), the National Highway Traffic Safety Administration (NHTSA), the European New Car Assessment Programme (EURO NCAP), the European Telecommunications Standards Institute (ETSI), United Nations (UN), and others. Vehicle safety is considered across various vehicle subsystems and application areas, including passive safety, active safety, EV safety, cybersecurity and data safety, V2X communication safety, perception and ADAS safety, functional safety, occupant \& VRU safety. Considering the scope of this paper, only the standards concerned with the following vehicle safety categories are discussed:
\begin{itemize}
    \item \textbf{Active safety:} Driver assistance and vehicle control systems
    \item \textbf{Cybersecurity and data safety:} cyber threats detection and protection
    \item \textbf{V2X communication safety:} Safe and secure V2X communication
    \item \textbf{Perception and ADAS safety:} Accuracy and reliability of sensor systems
    \item \textbf{Occupant and VRU safety:} Protection of occupants and vulnerable road users (VRUs)
\end{itemize}

Two of the most important standards in this area are ISO 26262~\cite{Gosavi2018ApplicationSurvey} and ISO 21448~\cite{Gotze2023Safety21448}. ISO 26262 defines a risk classification system known as the Automotive Safety Integrity Level (ASIL) for the functional safety of vehicles. ASIL is determined based on hazard analysis and risk assessment of a vehicle operating scenario. Factors such as severity, exposure, and controllability of the vehicle contribute to the ASIL estimation. ISO 26262 standard ensures that automotive systems are designed and built to prevent and handle failures within the functional safety framework of a vehicle. Functional safety for GNSS positioning systems were studied in~\cite{Pisoni2019GNSSVehicle, Infante2024DemonstrationVehicles} for autonomous land vehicles. Existing functional safety frameworks do not provide concrete information and guidance or evaluating Artifical Intelligence (AI) based systems, which are often deployed in autonomous vehicles. Diemert et al. \cite{Diemert2023SafetyIntelligence} proposed an AI based Safety Integirty Level (AI-SIL) framework which determines the complexity of an AI system and combines it with the existing 'Level of Rigor' approaches. The advancements in safety standards for machine learning-based road vehicle functions are reviewed in \cite{Burton2024NavigatingFunctions}.

However, with the emergence of newer sensor technologies, advanced driver functionalities are hugely dependent on cameras,  LiDARs, radars, and other sensors, to understand its surrounding environment. A vehicle can encounter sensor faults, and malfunctions due to adverse weather conditions or dynamic environment factors beyond the operational design domain of the safety framework. Sensor data are accessed to monitor the overall system health and the environment through a method called SOTIF (Safety of the Intended Functionality) governed by the ISO 21488 standard. ISO 21448 addresses safety concerns that arise from the intended functionality of the system, particularly in scenarios where the system is functioning as designed but still may pose safety risks. It is especially relevant for systems that rely on complex sensors, algorithms, and machine learning, like those used in automated driving. Challenges in Validating SOTIF are presented in \cite{Krishnan2022ValidationRisks/Hazards}, with particular attention to addressing false positives and false negatives for different levels of autonomy. With increasing levels of autonomy (level-3 and higher), scenario-based validation becomes important. However, it is not entirely possible to test all the scenarios with possible triggering conditions. Jimenez et al. presented a perception performance insufficiency approach \cite{Jimenez2024SafetyInjection} to validate SOTIF in different scenarios. Huang et al. \cite{Huang2021AEvents} proposed a systematic identification framework for triggering events consisting of system limitations and human errors.

In addition to ISO 21448 and ISO 26262 standards, there exists other ISO and UN regulatory standards focusing on active safety and ADAS systems. UN R79 and UN R152 respectively, oversee the steering control and automatic emergency brake systems (AEBS) requirements~\cite{AutomotiveAssociation} to enable functionalities such as lane-keeping assist and crash avoidance. In the realm of automated vehicles, ISO 23150 \cite{Han2023DesignSoftware} provides the logical interface for data communication between sensors and the data fusion unit for automated driving functions. ISO 16505~\cite{ISOProcedures} provides the minimum requirements for safety and performance of camera monitoring systems (CMS) in road vehicles. It offers enhanced visual information on the vehicle's surroundings, assisting the driver by improving visibility. Another important automotive standard that focuses on cybersecurity is ISO/SAE 21434~\cite{Li2024ComplyingVehicle} which defines common cybersecurity practices and provides a common language for risk communication and management.

\begin{table}[ht]
\centering
\caption{Comparison between ITS-G5 and C-V2X}
\begin{tabular}{|>{\centering\arraybackslash}m{2cm}|>{\centering\arraybackslash}m{2.5cm}|>{\centering\arraybackslash}m{3cm}|}
\hline
\textbf{} & \textbf{ITS-G5} & \textbf{C-V2X} \\
\hline
\textbf{Technology} & IEEE 802.11p & 3GPP (Cellular-based) \\
\hline
\textbf{Modes} & Direct communication only & Direct Mode (PC5) and Network Mode (Uu) \\
\hline
\textbf{Latency} & 1–2 ms & 1–2 ms in PC5; higher in Uu mode \\
\hline
\textbf{Range} & 300–500 meters & 300–500 meters (PC5); broader in Uu \\
\hline
\textbf{Infrastructure} & Requires RSUs & Uses existing cellular infrastructure \\
\hline
\textbf{Maturity} & Widely deployed in Europe & Newer, gaining support in North America and China \\
\hline
\textbf{Future-readiness} & Limited scalability & Highly scalable with 5G \\
\hline
\textbf{Regional Adoption} & Europe & North America, China \\
\hline
\end{tabular}
\label{table:ITS_G5_vs_C_V2X}
\end{table}

Apart from the onboard positioning and perception sensors, the vehicles can communicate information with surrounding vehicles and infrastructure through V2V, V2I, and V2X communication. Standards such as ISO 20077, ISO 20078, and ISO 21217 are essential in the context of connected and cooperative vehicles~\cite{Dominguez2019ReviewTime}, focusing on data exchange, security, and system architecture aspects. There are mainly two V2X technologies - namely ITS-G5 and Cellular-V2X (C-V2X)~\cite{Turley2018C-ITS:G5, Karoui2020PerformanceScenarios}. ITS-G5 is a short, direct communication protocol based on the IEEE 802.11p standard. It used the 5.9 GHz frequency, for low-latency, real-time communication. However, the C-V2X technology is a cellular-based technology developed by the 3rd Generation Partnership Project (3GPP). It operates in two modes: (a) Direct Mode (PC5), for direct communication between vehicles and infrastructures using the 5.9 GHz band without relying on cellular networks, and (b) Network Mode (Uu), which uses cellular networks (4G-LTE and preferably 5G) to provide long-range communication enabling sharing of traffic data via cloud sharing. A comparison between ITS-G5 and C-V2X technology is presented in TABLE~\ref{table:ITS_G5_vs_C_V2X}.

\begin{table*}[ht]
\centering
\caption{Summary of C-ITS Projects and Deployments In Europe}
\label{table:etsi_implementation}
\begin{tabular}{|p{2.5cm}|p{3cm}|p{3cm}|p{3cm}|p{2cm}|p{1.5cm}|}
\hline
\textbf{Project} & \textbf{Stakeholders} & \textbf{Key Applications} & \textbf{ETSI ITS-G5 Messages} & \textbf{Deployment Stage} & \textbf{Year} \\
\hline
DRIVE C2X~\cite{Schulze2014DRIVEReport} & DAIMLER, CRF, BMW, RENAULT, etc. & Emergency vehicle warning, In-vehicle signage, obstacle warning & CAM, DENM & Cross-border tests & 2011-2014 \\
\hline
Car2X~\cite{Schaal2012Car2x-FromProjects} & Volkswagen, BMW, VECTOR, etc. & Intersection movement assist, Road work alert, Platooning & CAM, DENM & Field trials, Pilot tests, Commercial deployements & 2013-ongoing \\
\hline
Scoop@F~\cite{SCOOPVehicle} & French ministry of transportation, Renault, PSA, METS, etc. & Intersection safety, Road work alert, onboard signalling & CAM, DENM & Field trials & 2014-2018 \\
\hline
NordicWay~\cite{Innamaa2024NordicWayReport} \newline - NordicWay 1 \newline - NordicWay 2 \newline - NordicWay 3 & EU, VOLVO, \newline SAFEROAD, SKANSKA, BM SYSTEM, etc. & Emergency vehicle warning, Road work alert, Traffic management & CAM, DENM, SPATEM, MAPEM & Cross-border deployment pilot, Interoperability tests & (2015-2023) \newline - 2015-2017 \newline - 2017-2020 \newline - 2019-2023 \\
\hline
5GAA~\cite{V2N2XExamples} & 3GPP, ETSI, OmniAir, 5G-IA/5GPPP, etc. & CACC, Automated valet parking, Platooning & DENM, IVIM, SPATEM,
MAPEM, SREM, SSEM, CAM, CPM & Field trials, Commercial deployments & 2016-ongoing \\
\hline
C-Roads platform~\cite{Documents:C-Roads} & EU member states, Renault, PSA, Garmin, VOLVO, Swarco, etc. & Road work alert, In-vehicle signage, Automated vehicle guidance, Collective perception & CAM, DENM, SPATEM, MAPEM, IVIM, SREM & Cross-border interoperability tests & 2016-ongoing\\
\hline
CONCORDA~\cite{Catana2021CONCORDAAutomation} & ERTICO, BOSCH, RENAULT, PSA, etc.  & High-density truck platooning & CAM, DENM & Cross-border interoperability tests & 2017-2021 \\
\hline
C-MobILE~\cite{Ferrandez2018ModellingStudy} & PTV group, Swarco, TomTom, CTAG, Applus IDIADA & Urban parking management, Dynamic eco-driving, CACC, Road work alerts & CAM, DENM, SPATEM, MAPEM, IVIM & Pilot tests, Urban \& regional deployments & 2017-2021\\
\hline
5G-Mobix~\cite{Trichias20195GMOBIXSpecifications} & Nokia, Valeo, Siemens, Ericsson, CTAG, etc. & Cooperative collision avoidance, Cloud-enabled platooning, Automated shuttle remote driving & CAM, DENM, CPM & Pre-deployment trials, Cross-border tests & 2018-2022 \\
\hline
5G-ROUTES~\cite{Rodriguez20235GApproach} & AIRBUS, ERICSSON, LMT, TELIA, CERTH, etc. & Cooperative collision avoidance, Vehicle platooning, state sharing & CAM, CPM, DENM, MAPEM, SPATEM & Cross-border tests & 2021-ongoing \\
\hline
\end{tabular}
\end{table*}

The ETSI ITS-G5 standards developed by the European Telecommunications Standards Institute (ETSI) are designed to enable reliable, real-time V2X communication, especially for critical cooperative safety applications like collision warning, emergency braking, and road hazard notifications. ETSI ITS-G5 has been widely adopted across Europe, primarily in EU-funded projects like SCOOP@F and C-Roads,
where countries collaborate on cross-border interoperability
and safety~\cite{SCOOPVehicle,Documents:C-Roads}. The implementation of ETSI ITS-G5 by various
automakers and research groups is provided in the TABLE IV. A few key ETSI ITS-G5 standards are as follows:\\

\begin{itemize}
    \itemsep 1mm
    \item \textbf{ETSI EN 302 637-2: Cooperative Awareness Messages (CAM)~\cite{2014ENService}}\\ The standard defines the periodic Cooperative Awareness Messages (CAM) that the vehicles broadcast to provide essential information such as position, speed, heading, etc. CAM Messages enhance situational awareness critical for collision avoidance, lane-keeping, and cooperative adaptive cruise control (CACC), where precise vehicle positions are required for safe and coordinated maneuvers. It is quite similar to the Basic Safety Messages (BSM) defined under SAE standards.
    
    \item \textbf{ETSI EN 302 637-3: Decentralized Environmental Notification Messages (DENM)~\cite{2019ENService}}\\ DENM messages improve real-time response to unpredictable events, allowing vehicles to adjust speed, change lanes, or even stop in advance of encountering hazards. This standard is vital for emergency electronic braking and real-time traffic updates, allowing vehicles to communicate situational hazards that can prevent collisions and protect vulnerable road users.
    
    \item \textbf{ETSI EN 302 665: ITS Communications Architecture~\cite{2010ENArchitecture}}\\ This standard specifies the architecture for ITS communications in a way that enables interoperability between different ITS stations, including vehicles, RSUs, and central management systems. In high-density traffic environments, this interoperability enables coordinated safety maneuvers and effective traffic management, essential for applications like lane merging and intersection management.
    
    \item \textbf{ETSI EN 302 571: Radio Spectrum Requirements for ITS~\cite{2017EN2014/53/EU}}\\ This standard regulates the use of the 5.9 GHz frequency band specifically for ITS applications, minimizing interference and ensuring secure communication for V2X applications. By dedicating the 5.9 GHz band to ITS, this standard helps prevent interference from non-ITS devices, preserving the reliability and integrity of critical safety messages.
    
    \item \textbf{ETSI TR 102 863: Local Dynamic Map (LDM) standardization~\cite{2011TRStandardization}}\\ This technical report provides standardization of LDM based on the requirements of various ITS applications. It acts as a central repository storing both static elements such as road traffic road signs and dynamic elements like moving vehicles. The LDM acquires data from various entities such as vehicles, on-board units, roadside units, ITS stations, etc., which is essential to deploy cooperative ITS applications.
\end{itemize}

\begin{table*}[!ht]
\centering
\caption{Summary of C-ITS Projects and Deployments In USA}
\label{tab:J2945_projects}
\begin{tabular}
{|p{3cm}|p{3cm}|p{3cm}|p{3cm}|p{2cm}|p{1.5cm}|}
\hline
\textbf{Project} & \textbf{Stakeholders} & \textbf{Key Applications} & \textbf{SAE J2735 Messages} & \textbf{Deployment Stage} & \textbf{Year} \\
\hline
Crash Avoidance Metrics Partnership (CAMP)~\cite{Goudy2022} & Ford, GM, FHWA, etc. & Traffic optimization, red light violation warning, queue advisory & BSM, SPaT, MAP & Field Testing, Interoperability Testing & 1999-ongoing\\
\hline
Safety Pilot Model Deployment (SPMD)~\cite{Gay2015SafetyActivities} & UMich, USDOT, ITS JPO, NGTSA, FHWA, etc. & Forward collision avoidance, intersection movement assist, left turn assist, blind spot warning & BSM, SPaT, TIM & Pilot Demonstrations, Interoperability Testing, Data collection & 2010-2014\\
\hline
Heavy truck CACC project~\cite{Shladover2018CooperativePlatooning} & FHWA \& Auburn university, Peloton technology, Meritor, Inc., etc. & Truck platooning, CACC & BSM & Field Testing & 2013-2019\\
\hline
Ann Arbor Connected Vehicle Test Environment (AACVTE)~\cite{Wang2020DataApplications} & UMTRI, USDOT, Savari, Denso, Cohda Wireless, etc. & \{Same as SPMD project\}& BSM, SPaT, MAP, TIM, PSM  & Large-scale Operational Deployment & 2015-2018\\
\hline
USDOT Connected Vehicle Pilots~\cite{USDOTProgram}\newline - New york city pilot\newline - Tampa (THEA) pilot\newline - Wyoming DOT pilot & USDOT, ITS JPO, NYCDOT, WYDOT, THEA, etc. & Intersection movement assist, forward collision avoidance, red light violation warning, probe enabled traffic monitoring &  BSM, SPaT, MAP, TIM, PSM & Pilot Deployment, Interoperability Testing & 2015-2022\\
\hline
SAFE SWARM~\cite{HondaSWARM} & Honda, Ohio DOT, Verizon, etc. & Vehicle-to-vehicle obstacle detection, lane speed monitoring, braking event communication & BSM, SPaT & Field Testing & 2017-ongoing\\
\hline
ITS-America national C-V2X deployment plan~\cite{ITSAmerica2023ITSCollaboration, US-DOTITSJointProgramOffice2024SavingContents} & USDOT, FHWA, FCC, NTIA, Automotive OEMs, State and local transportation agencies, etc. & Intersection movement assist, left-turn assist, eco-driving, traffic signal priority, etc. & BSM, SPaT, MAP, TIM, PSM, SRM, SSM, RSM & Planning \& Deployment & 2024-ongoing\\
\hline
\end{tabular}
\end{table*}

Unlike ETSI ITS-G5, standards SAE J2735 and SAE J2945 designed by the Society of Automotive Engineers (SAE) primarily focuses on the C-V2X, though it is also applicable to ITS-G5. J2735 defines message sets, data frame and data elements used in V2X communication. The standardization of messages and its structure facilitates interoperability between V2X communication devices. Few of the key J2735 messages include Basic Safety Message (BSM), Singal Phase and Timing (SPaT), Map message (MAP), Traveler Information Message (TIM) and many more. A list of experiments demonstrating the use of the SAE J2735 messages is mentioned in TABLE~\ref{tab:J2945_projects}. J2945 defines performance requirements and guidelines for V2X safety applications, focusing on ensuring interoperability, safety, and reliability in vehicle-to-everything (V2X) communication. Some of the important J2945 standards are summarized as follows:\\

\begin{itemize}
    \itemsep1mm
    \item \textbf{SAE J2945/1: On-Board System Requirements for V2V Safety Communications~\cite{J2945/1_202004}} \\
    Defines the minimum performance requirements for \textit{Basic Safety Messages (BSM)} that vehicles broadcast in Vehicle-to-Vehicle (V2V) safety applications. BSMs contain essential vehicle data, such as position, speed, heading, brake status, and size. These messages enable applications like forward collision warning, emergency braking, and lane change warnings. The standard specifies the message frequency (typically 10 Hz) and message structure to ensure timely and accurate data exchange, critical for collision avoidance.

    \item \textbf{SAE J2945/2: DSRC Performance Requirements for V2V Safety Awareness~\cite{J2945/2_201810}} \\
    Establishes the performance requirements for \textit{Vehicle-to-Vehicle (V2V)} safety awareness applications, particularly using Dedicated Short-Range Communications (DSRC). The standard focuses on V2V applications that involve infrastructure components such as roadside units (RSUs) and traffic signals. The use cases covered in this document include (a) Emergency Vehicle Alert (EVA), (b) Roadside Alert (RA), (c) Safety Awareness - Object (SAW - O), and (d) Safety  Awareness - Adverse road conditions (A).

    \item \textbf{SAE J2945/4: Road Safety Applications~\cite{J2945/4_202305}} \\
    The standard specifies message set requirements for different roadside safety applications which include: (a) Curve Speed Warning (CSW), (b) Reduced Speed Zone Warning (RSZW), (c) Lane Closure Warning (LCW), (d) Dynamic Traveler Information (DTI), and (e) Incident Information (INC). The document also presents the Road Safety Message (RCM) and the application of I2V communication for roadside safety. The message set and structure definition help achieve interoperability between I2V communication devices.

    \item \textbf{SAE J2945/7: High-Precision Positioning for V2X Systems~\cite{J2945/7_202308}} \\
    J2945/7 addresses new use cases that can benefit from increased precision and integrity in positioning systems. High-Precision Positioning Systems (HPPS) equipped V2X devices would allow to assess the quality of data being shared among road agents, especially vehicle state information such as position, and speed, which can be integrated with on-board positioning system to enable automated driving. Applications such as Emergency Brake Light (EEBL), Pedestrian cooperative detection, Cooperative automation platooning can be enhanced through the use of HPPS.

    \item \textbf{SAE J2945/9: Vulnerable Road User (VRU) Safety~\cite{J2945/9_201703}} \\
    Defines requirements for communicating Personal Safety Message (PSM) between vehicles and \textit{Vulnerable Road Users (VRUs)}, such as pedestrians, bicyclists, and public safety personnel. This standard enables Vehicle-to-Pedestrian (V2P) applications where VRUs equipped with communication devices can broadcast their position and movement status to nearby vehicles equipped with Dedicated Short Range Communication (DSRC) devices. The document emphasizes the use of DRSC for broadcasting the PSM, although different wireless means may be used. J2945/9 sets criteria for alerting drivers about VRUs in potential collision paths, especially in urban environments with mixed traffic.

\end{itemize}

\begin{figure*}[ht]
        \centering        
        \includegraphics[width=\textwidth]{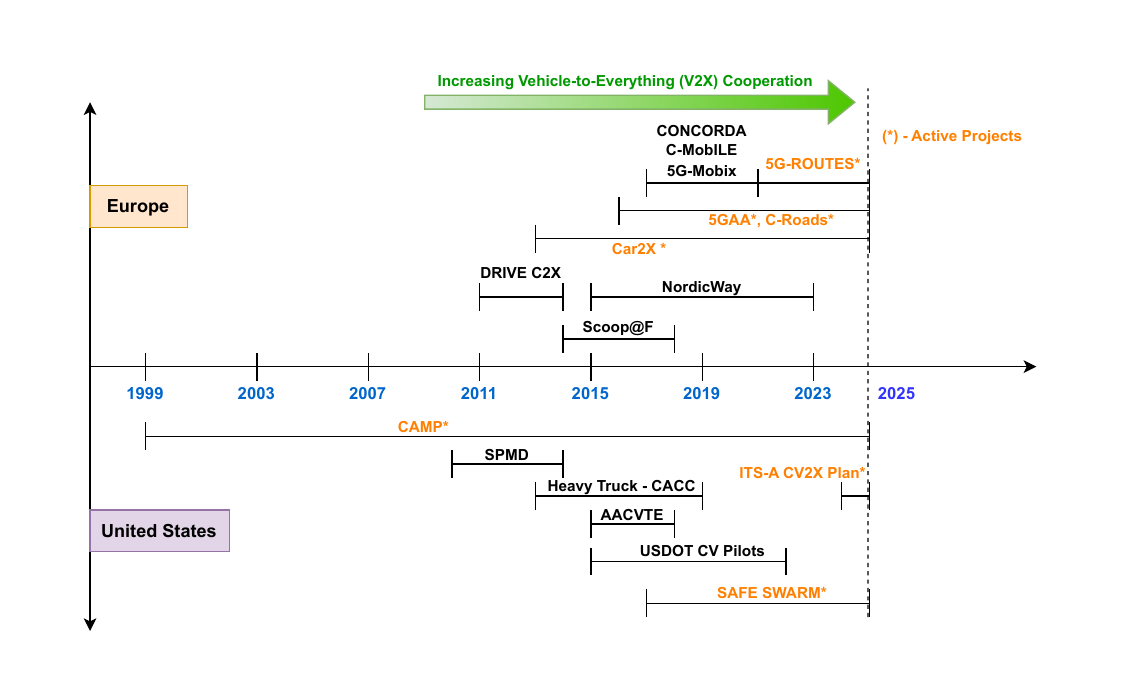}
        \caption{Timeline of major Cooperative ITS and Connected Vehicle pilot projects in Europe (top row) and the United States (bottom row) from 1999 through 2025. Projects marked with an asterisk (*) are currently active.}
        \label{fig: ITS_projects}
\end{figure*}\

Today, many vehicles are already equipped with Advanced Driver-Assistance System (ADAS) and are able to perform level 1-2 autonomy tasks such as Forward Collision Warning (FCW), Lane Keeping Assistance (LKA), Automatic Emergency Braking (AEB) and many other functions (refer to SAE J3016~\cite{J3016_202104}). On the other hand, Level 3-5 vehicles (e.g. Waymo) equipped with Automated Driving Systems (ADS) are able to perform Dynamic Driving Tasks (DDT) and Object and Event Detection and Response (OEDR) with extended Operational Design Domain (ODD)(refer to SAE J3216~\cite{J3216_202107}). 

As per the SAE J3216, Cooperative Driving Automation (CDA), cooperation between two or more road agents through Machine-to-Machine (M2M) communication can further enhance driver safety and improve traffic flow. It can enhance the performance of DDTs while the driving autonomous features are engaged in a vehicle. The Federal Highway Administration (FHWA) launched the open-source software - CARMA~\cite{Lou2022FHWAFHWA} in 2013 to demonstrate CDA and test the V2X capabilities and algorithms. Other notable open-source CDA frameworks include OpenCDA~\cite{Xu2021OpenCDA:Co-Simulation} and CoDrivingLLM~\cite{Fang2024TowardsFramework}. OpenCDA focuses on developing a unified open-source CDA framework, while CoDrivingLLM leverages Large Language Models (LLMs) for decision-making in critical traffic scenarios.

SAE J3216 presents the taxonomy and definitions related to CDA, whereas other SAE standards focus on Concept of Operations (CONOPS), use cases, message flows, and performance requirements for specific CDA applications. Some of these standards are mentioned below:
\begin{itemize}
    \item \textbf{SAE J3186:} Application protocol and requirements for maneuver sharing and coordinating service (refer to~\cite{J3186_202303});
    \item \textbf{SAE J3224:} V2X sensor-sharing for cooperative and automated driving (refer to~\cite{J3224_202208});
    \item \textbf{SAE J3282:} CDA Feature - Cooperative permissive left turn across opposing traffic with infrastructure guidance (refer to~\cite{J3282_202406});
    \item \textbf{SAE J3256:} CDA Feature - Infrastructure-based prescriptive cooperative merge (refer to~\cite{J3256_202403}); and
    \item \textbf{SAE J3251:} CDA Feature - Perception status sharing for occluded pedestrian collision avoidance (refer to~\cite{J3251_202308}).
\end{itemize}

\section{Real-world Cooperative Driving Automation Experiments and Public-Datasets}
\label{sec: v2x datasets}

The development and testing of integrity solutions for Cooperative Driving Automation (CDA) systems heavily rely on high-quality public datasets, as they provide foundational sensor data, environmental context, and multi-agent interactions for experimentation and benchmarking. Datasets~\cite{Liu2024AOutlook} like KITTI, Waymo, and nuScenes have set industry standards by providing robust datasets for single-agent autonomous driving. These datasets include comprehensive sensor logs from onboard GNSS, IMUs, cameras, LiDARs, and radars, but focus predominantly on scenarios without active vehicle-to-vehicle (V2V) or vehicle-to-infrastructure (V2I) communications. As such, they primarily support perception tasks in isolated, single-agent settings and are less suitable for cooperative tasks central to CDA, such as shared situational awareness or collaborative decision-making.

Recognizing the limitations of single-agent datasets, new V2X datasets have been introduced in the last three years to capture multi-agent cooperative interactions. These datasets feature sensor logs from both onboard and infrastructure sensors, offering data in vehicle-to-vehicle and vehicle-to-infrastructure (V2V and V2I) settings. These data logs enable research on perception and decision-making for multi-agent interactions, enhancing autonomous systems with cooperative awareness and more accurate situational assessments.

The V2X datasets (see TABLE~\ref{tab:v2x_datasets}) serve various purposes such as 3D detection, tracking, trajectory prediction, and communication performance evaluation. V2X-Sim~\cite{Li2022V2X-Sim:Driving} data were collected from vehicles and infrastructure sensors in a SUMO and CARLA-built co-simulation environment. V2V4Real~\cite{Xu2023V2V4Real:Perception} focuses on data collected from two test vehicles in urban and highway conditions involving only V2V scenarios. However, data sets such as DAIR-V2X~\cite{Yu2022DAIR-V2X:Detection}, TUMTraf-V2X~\cite{Zimmer2024TUMTrafDataset}, and V2X-Real~\cite{Xiang2024V2X-Real:Perception} provide data that contain both V2V and V2I scenarios. V2X-Seq~\cite{Yu2023V2X-Seq:Forecasting}, a subset of DAIR-V2X provides sequential trajectory data at intersections for trajectory forecasting. The aforementioned datasets are collected mostly at intersections or in urban conditions. The H-V2X dataset consisting of the V2I scenario collected in China spans more than 100 km of highway roads with cameras and radars mounted on overhead masts. 

\newcommand{\GPS}{G} % GPS symbol
\newcommand{\CAM}{C} % Camera symbol
\newcommand{\LIDAR}{L} % LiDAR symbol
\newcommand{\RAD}{R} % Radar symbol
\newcommand{\IMU}{I} % Radar symbol
\newcommand{\VtoX}{V2X}

\begin{table*}[ht]
\centering
\caption{Summary of V2X Datasets}
\label{tab:v2x_datasets}
\begin{tabular}{|p{1.9cm}|p{0.6cm}|p{2cm}|p{2.5cm}|p{2cm}|p{1.9cm}|p{1.6cm}|p{2.7cm}|}
\hline
\textbf{Dataset} & \textbf{Year} & \textbf{Source} & \textbf{Test Infrastructure} & \textbf{Sensors /\newline Information} & \textbf{Traffic \newline Environment} & \textbf{V2X Comm. \newline Protocol} & \textbf{Use Cases}\\ 
\hline
V2X-Sim~\cite{Li2022V2X-Sim:Driving} & 2022 & \textit{In-Simulation: } \newline SUMO-CARLA &  Multiple test vehicles / infrastructure  & \textit{OB}: \GPS, \IMU, \LIDAR, \CAM \newline \textit{RS}: \LIDAR, \CAM &  Intersection & N.A & 3D object detection, tracking, segmentation\\ 
\hline
DAIR-V2X~\cite{Yu2022DAIR-V2X:Detection} & 2022 & Beijing, China & Multiple test vehicles, 28 smart-intersection & \textit{OB}: \GPS, \IMU, \LIDAR, \CAM \newline \textit{RS}: \LIDAR, \CAM & Intersection & N.A & 3D object detection  \\
\hline
V2X-Seq~\cite{Yu2023V2X-Seq:Forecasting} & 2023 & Beijing, China & Multiple test vehicles, 28 smart-intersection & \textit{OB}: \GPS, \IMU, \LIDAR, \CAM \newline \textit{RS}: \LIDAR, \CAM & Intersection & N.A & 3D tracking, Trajectory forecasting \\
\hline
V2V4Real~\cite{Xu2023V2V4Real:Perception} & 2023 & Columbus, USA & 2 test vehicles & \textit{OB}: \GPS, \IMU, \LIDAR, \CAM & Highway, Urban & N.A & 3D object detection, tracking, transfer learning\\
\hline
Berlin-V2X~\cite{Hernangomez2023BerlinTechnologies} & 2023 & Berlin, Germany & 4 test vehicles & Cellular, Sidelink, GPS  & Highway, Urban & LTE, SDR &  multi-RAT QoS prediction, transfer learning \\
\hline
TUMTraf-V2X~\cite{Zimmer2024TUMTrafDataset} & 2024 & Munich, Germany & 1 test vehicle, 1 smart-intersection & \textit{OB}: \GPS, \IMU, \LIDAR, \CAM \newline \textit{RS}: \LIDAR, \CAM  & Intersection & N.A & 3D object detection, tracking \\
\hline
V2X-Real~\cite{Xiang2024V2X-Real:Perception} & 2024 & L.A, USA & 2 test vehicles, 2 smart-intersection & \textit{OB}: \GPS, \IMU, \LIDAR, \CAM \newline \textit{RS}: \GPS, \LIDAR, \CAM  & Intersection, \newline Urban corridor & N.A & 3D object detection \\
\hline
OpenAD~\cite{Xiang2024V2X-Real:Perception} & 2024 & L.A, USA & 2 test vehicles, 2 smart-intersection & \textit{OB}: \GPS, \IMU, \LIDAR, \CAM \newline \textit{RS}: \GPS, \LIDAR, \CAM  & Intersection, \newline Urban corridor & N.A & 3D object detection \\
\hline
H-V2X~\cite{Liu2025H-V2X:Perception} & 2024 & China & Smart-infrastructure spanning 100 Km & \textit{RS}: \CAM, \RAD  & Highway & N.A & BEV detection \& tracking, trajectory forecasting \\
\hline
V2AIX~\cite{Kueppers2024V2AIX:Traffic} & 2024 & Germany & V2X-enabled sensor-rich test vehicles \& \newline infrastructure & \textit{OB}: \GPS, \IMU, \LIDAR, \CAM, \VtoX \newline \textit{RS}: \LIDAR, \CAM, \VtoX & Highway, Urban, Rural  &  ETSI ITS-G5 & Vehicle Localization, V2X 
message \newline  standardization, CAV penetration analysis  \\
\hline
TiHAN-V2X~\cite{TIHAN-IITH}& 2024 & Hyderabad, India & 2 test vehicles, 1 smart-intersection & \textit{OB}: \GPS, \VtoX \newline \textit{RS}: \VtoX & Highway, Urban, Rural & 5G-sidelink & Communication performance analysis - latency, path loss, throughput, SNR, etc. \\
\hline
% Continue adding rows as necessary
\end{tabular}
\label{table:dataset_summary}
\vspace{0.3cm} % Space between table and legend

\begin{tabular}{l}
    \textbf{Legend:} \\
    \GPS = GNSS, \IMU = IMU, \CAM = Camera, \LIDAR = LiDAR, \RAD = Radar, \VtoX = V2X communication module,\\
    OB: Onboard, RS: Roadside
\end{tabular}
\end{table*}

\begin{figure}[ht]
        \centering        
        \includegraphics[width=\linewidth]{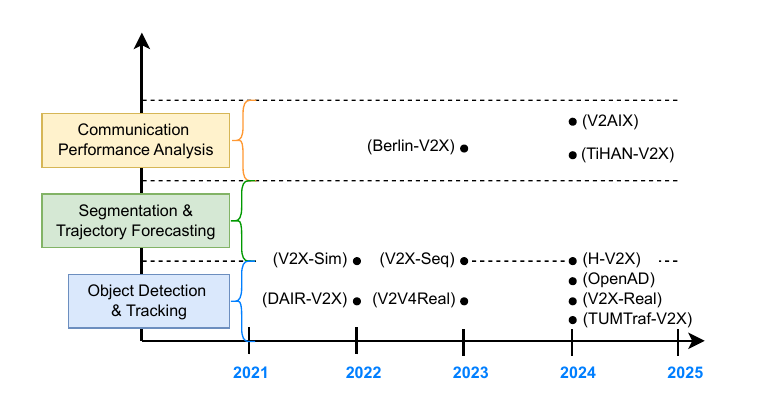}
        \caption{Evolution of V2X datasets from object detection \& tracking to segmentation \& trajectory forecasting, with a growing number of recent releases emphasizing CAV communication performance.}
        \label{fig: cv2x_datasets}
\end{figure}

These V2X datasets mostly focus on cooperative perception and sensor fusion problems and offer a benchmark for 3D detection, tracking, and trajectory prediction analyses. However, the dataset is collected locally at source i.e. vehicle, infrastructure, and doesn't involve active V2X communication among the agents. The communication constraints such as bandwidth, latency, packet drops, etc., are imposed in post-processing analysis. Recently released datasets such as Berlin-V2X, V2AIX, and TiHAN-V2X, involve active communication among the agents and have collected data for evaluating real-time communication performance, as shown in fig\ref{fig: cv2x_datasets}. Berlin-V2X~\cite{Hernangomez2023BerlinTechnologies} and TIHAN-V2X~\cite{TIHAN-IITH} primarily provide information on cellular and sidelink communication in V2X scenarios that can help evaluate Quality of Service (QoS) Prediction and ML algorithms focusing on optimizing the communication channel. V2AIX dataset~\cite{Kueppers2024V2AIX:Traffic}, released in 2024, consists of both cooperative perception data (V2V \& V2I) and ETSI ITS-G5 V2X messages such as CAM, DENM, MAPEM, and SPATEM. Such datasets are crucial for analyzing the effects of CAV penetration rate on various applications under real-world communication constraints.

Despite recent advances, CDA research faces challenges due to the scarcity of datasets incorporating adversarial conditions, such as spoofed GNSS signals or real-time V2X messaging data under unreliable communication networks. Such scenarios are critical for testing system robustness against location spoofing, signal jamming, and data latency—factors that can severely impact automated vehicle safety and decision-making in the real world. Future datasets could improve by including edge cases, various weather and lighting conditions, and data with sporadic GNSS availability. Expanding upon these aspects will enable more resilient CDA systems, capable of handling a wider variety of environments and potential adversarial events.

\section{Research Gaps and Future Directions}
\label{sec:gaps_n_future_dir}
In this study, we reviewed the existing work on integrity monitoring for connected and automated vehicles. This section describes the implications of our findings and discusses potential limitations and possibilities for future research. 

\begin{itemize}
    \item \textbf{Classical GNSS-based IMs: }The majority of the research to date has been focused on IM methods for GNSS-based systems. This is primarily due to the ubiquitous nature of GNSS positioning systems and well-defined RNPs for GNSS-based navigation. Recently, researchers have started exploring integrity solutions for alternate GNSS-based positioning systems involving Signals of Opportunity, and communication techniques such as UWB and 5G-sidelink. \newline
    
    \item \textbf{IM for onboard perception systems: }RNPs for navigation systems involving various other onboard sensors, such as camera, LiDAR, and radar are being developed for intelligent CAV applications. However, the research can be quickly brought to pace by leveraging Hardware-in-the-Loop (HiL) setups to test different sensor modalities, driving, and environment conditions that would be otherwise difficult to test in real-world conditions. Corner cases and external conditions such as weather, and communication constraints could be imposed in a simulation environment to evaluate and validate IM methods. \newline
    
    \item \textbf{IM for V2X-based perception systems: }There is a lack of IM methods to evaluate integrity risks in CAV applications that involve the sharing of perception data between vehicles. There is a wide variety of publicly available V2X datasets (see TABLE~\ref{tab:v2x_datasets}), but they are mostly used for cooperative perception tasks. Thus far, only a small amount of attention has been paid to evaluating integrity measures using these datasets. This could help quantify RNP parameters and provide new insights for developing new IM techniques for CAV applications in various traffic conditions. \newline
    
    \item \textbf{Sensitivity analyses: }Furthermore, the performance of a CAV application on a vehicle depends on the type of vehicle, sensor modality, communication performance, and traffic environment. Hence, thorough sensitivity analyses are required to establish the RNPs for CAV applications in mixed-traffic situations. This could be achieved by building virtual traffic environments in simulation platforms e.g. CARLA, with various road agents and sensor modalities to evaluate critical traffic conditions. \newline
    
    \item \textbf{Integrity for AI systems: }In the last decade, Artificial Intelligence (AI) algorithms e.g. Machine Learning (ML) have become ubiquitous in the field of ITS. The emergence of deep neural networks and improvements in computational capabilities have made it possible to run real-time perception, planning, and control tasks. In addition, with the advent of generative AI and Large Language Models (LLMs), there is a buzz in the research community on the possible use of these models to handle complex perception tasks. However, few researchers have looked into the error quantification and reliability of such models. This could be a challenging yet interesting research topic in the future, especially for end-to-end deep neural net-based methods.
    \newline
    
    \item \textbf{Adherence to automotive standards: }Researchers and automotive OEMs should form a consensus on defining the RNPs that adhere to the existing automotive safety standards. This will help to encourage the early adoption of the developed FDI and IM methods in manufactured vehicles.
    
\end{itemize}

\section{Conclusions}
\label{sec: conclusions}

This paper reviews existing Integrity Monitoring (IM) methods used in vehicle navigation and advocates the need for the development of new IM techniques in the field of Connected and Automated Vehicle (CAV) research. The consideration of positioning uncertainties in past CAV-related experiments is discussed followed by the definition of minimum Required Navigation Performance (RNP) criteria used in IM. It is found that most of the existing IM research focuses on onboard GNSS-based positioning systems. Although there exists a plethora of cooperative IM methods, the external sensor information such as Signals of Opportunity, DSRC, etc., are only used to eliminate the errors (multipath, NLOS) in GNSS signals. The integrity evaluation for cooperative positioning solutions involving V2V and V2I-based perception sensors are almost non-existent, even if various perception data-based V2X datasets are available in abundance. This might be due to the primary use cases of the aforementioned V2X datasets being 3D detection, tracking, and trajectory forecasting. Automotive safety standards and protocols are reviewed to gain insight into the standardization of V2X messages. These standards can be adopted by research communities for the development of new CA-centric IM frameworks. In future work, sensitivity analyses will be performed in simulation environments to identify safety-critical traffic maneuvers in various CAV applications and define a benchmark RNP chart that can be referred to by future researchers.

\section*{Acknowledgments}
This work was funded, partially or entirely, by a grant from the US Department of Transportation’s University Transportation Centers Program through the Center for Assured \& Resilient Navigation in Advanced Transportation Systems (CARNATIONS). The contents of this paper reflect only the views of the authors, who are responsible for the facts and the accuracy of the data presented.

\bibliographystyle{IEEEtran}
\bibliography{reference.bib}

\begin{IEEEbiography}[{\includegraphics[width=1in,height=1.25in,clip,keepaspectratio]{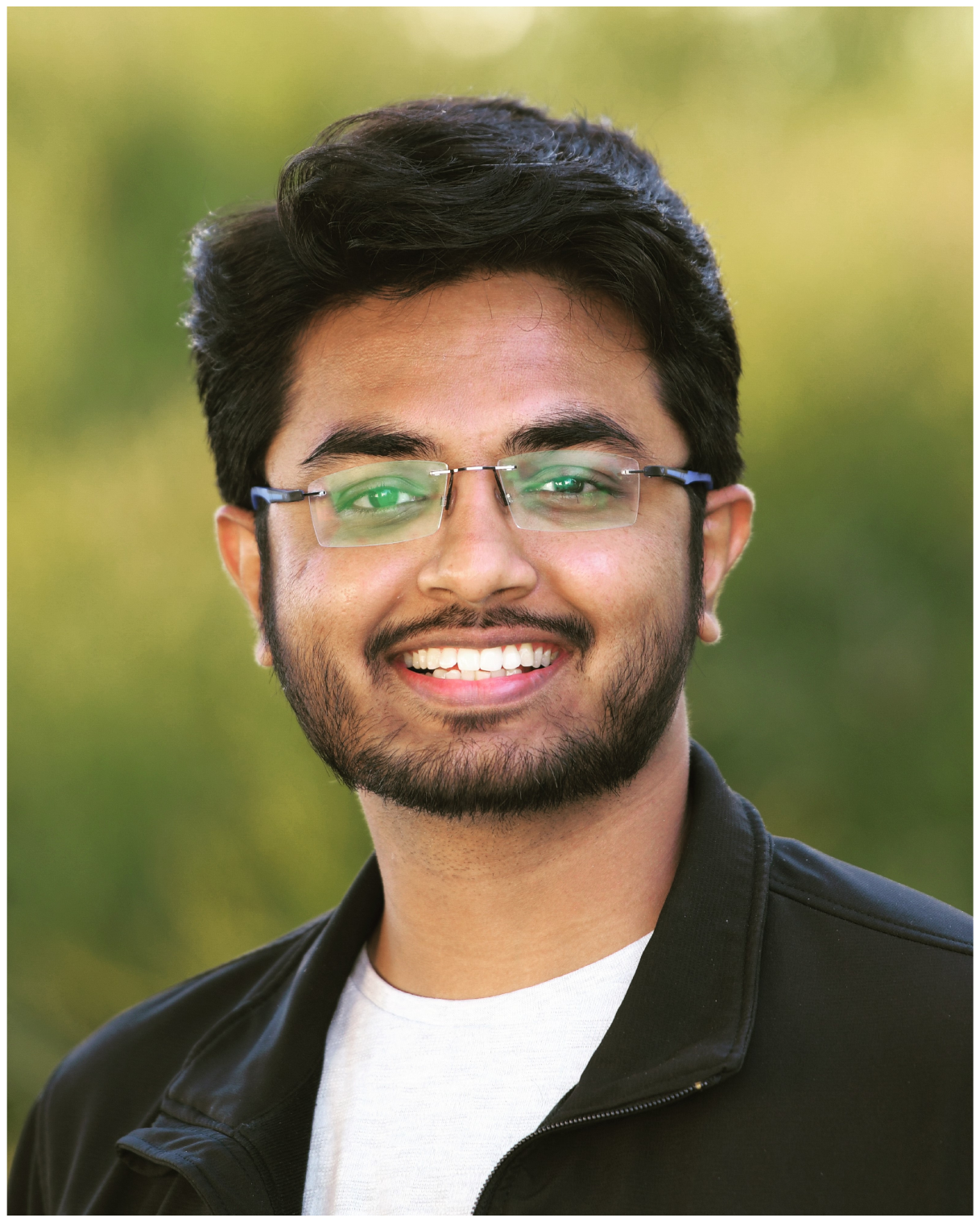}}]{Saswat Priyadarshi Nayak}
received the B. Tech degree in Electrical Engineering from the National Institute of Technology Rourkela, India in 2018. He served as a Project Associate at the Department of Aerospace Engineering, Indian Institute of Technology Kanpur, India 2018-19. He is currently pursuing a Ph.D. degree at the Center of Environmental Research and Technology (CE-CERT), University of California Riverside, USA. He is one of the recipients of the 2024 ITS California (ITS CA) and California Transportation Foundation (CTF) Scholarship. His main research interests include vehicle positioning and localization in mixed traffic scenarios, multi-sensor fusion, and connected vehicle applications.
\end{IEEEbiography}

\begin{IEEEbiography}[{\includegraphics[width=1in,height=1.25in,clip,keepaspectratio]{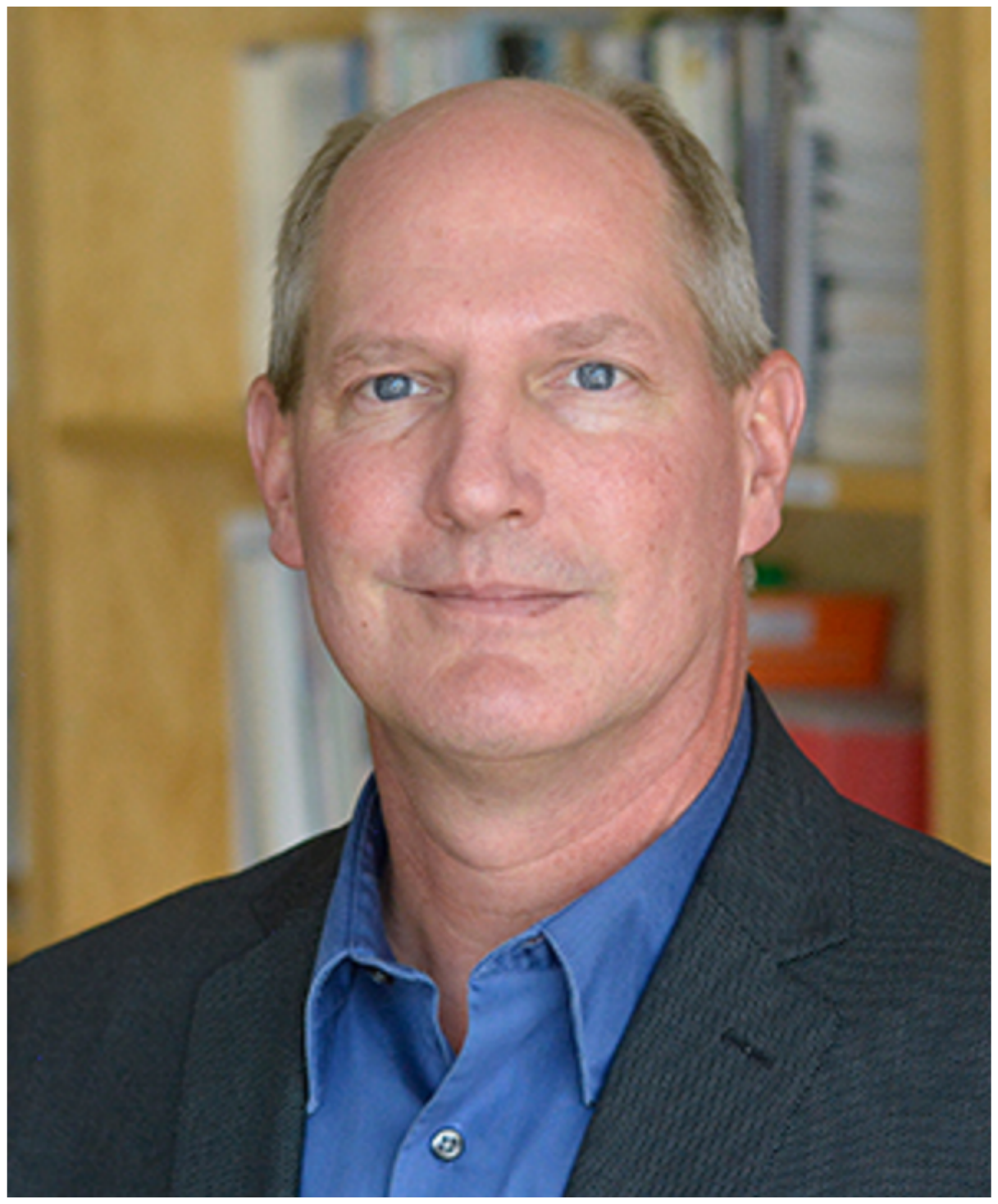}}]{Matthew J. Barth}
(Fellow, IEEE) received the M.S. and Ph.D degree in electrical and computer engineering from the University of California at Santa Barbara, in 1985 and 1990, respectively. He is currently the Hays Families Professor in the College of Engineering, University of California at Riverside, USA. He also served as the Director of the Center for Environmental Research and Technology. His current research interests include ITS and the environment, transportation/emissions modeling, vehicle activity analysis, advanced navigation techniques, electric vehicle technology, and advanced sensing and control. Dr. Barth has been active in the IEEE Intelligent Transportation System Society for many years, serving as a Senior Editor for both the Transactions of ITS and the Transactions on Intelligent Vehicles. He served as the IEEE ITSS President for 2014 and 2015 and is currently the IEEE ITSS Vice President of Education.
\end{IEEEbiography}

\vfill

\end{document}